\documentclass[pdflatex,sn-mathphys-num]{sn-jnl}

\usepackage{graphicx}%
\usepackage{multirow}%
\usepackage{amsmath,amssymb,amsfonts}%
\usepackage{amsthm}%
\usepackage{mathrsfs}%
\usepackage[title]{appendix}%
\usepackage{xcolor}%
\usepackage{textcomp}%
\usepackage{manyfoot}%
\usepackage{booktabs}%
\usepackage{algorithm}%
\usepackage{algorithmicx}%
\usepackage{algpseudocode}%
\usepackage{listings}%
\usepackage{xcolor}
\newcommand{\rev}[1]{\textcolor{black}{#1}}

\theoremstyle{thmstyleone}%
\theoremstyle{thmstyletwo}%

\theoremstyle{thmstylethree}%

\begin{document}
\thispagestyle{empty}
\noindent Notice: This manuscript has been authored by UT-Battelle, LLC, under Contract No. DE-AC0500OR22725 with the U.S. Department of Energy. The United States Government retains and the publisher, by accepting the article for publication, acknowledges that the United States Government retains a non-exclusive, paid-up, irrevocable, world-wide license to publish or reproduce the published form of this manuscript, or allow others to do so, for the United States Government purposes. The Department of Energy will provide public access to these results of federally sponsored research in accordance with the DOE Public Access Plan (\url{http://energy.gov/downloads/doe-public-access-plan}).
\clearpage
\title[Article Title]{Quality-Controlled Active Learning via Gaussian Processes for Robust Structure--Property Learning in Autonomous Microscopy}








\author[1]{\fnm{Jawad} \sur{Chowdhury}}\email{chowdhurym1@ornl.gov}
\author[1]{\fnm{Ganesh} \sur{Narasimha}}\email{narasimhag@ornl.gov}
\author[2]{\fnm{Jan-Chi} \sur{Yang}}\email{janchiyang@phys.ncku.edu.tw}
\author[3]{\fnm{Hiroshi} \sur{Funakubo}}\email{funakubo.h.31df@m.isct.ac.jp}
\author[4]{\fnm{Yoshitaka} \sur{Ehara}}\email{yoshieha@nda.ac.jp}
\author[1]{\fnm{Yongtao} \sur{Liu}}\email{liuy3@ornl.gov}
\author*[1]{\fnm{Rama} \sur{Vasudevan}}\email{vasudevanrk@ornl.gov}

\affil[1]{\orgdiv{Center for Nanophase Materials Sciences}, \orgname{Oak Ridge National Laboratory}, 
\orgaddress{\city{Oak Ridge}, \state{TN}, \postcode{37831}, \country{USA}}}

\affil[2]{\orgdiv{Department of Physics}, \orgname{National Cheng Kung University}, 
\orgaddress{\city{Tainan}, \postcode{70101}, \country{Taiwan}}}

\affil[3]{\orgdiv{Department of Materials Science and Engineering}, \orgname{Institute of Science Tokyo}, 
\orgaddress{\city{Yokohama}, \postcode{226-8502}, \country{Japan}}}

\affil[4]{\orgdiv{Department of Communications Engineering}, \orgname{National Defense Academy}, 
\orgaddress{\city{Yokosuka}, \postcode{239-8686}, \country{Japan}}}
\abstract{Autonomous experimental systems are increasingly used in materials research to accelerate scientific discovery, but their performance is often limited by low-quality, noisy data. This issue is especially problematic in data intensive structure--property learning tasks such as Image-to-Spectrum (Im2Spec) and Spectrum-to-Image (Spec2Im) translations, where standard active learning strategies can mistakenly prioritize poor quality measurements. We introduce a gated active learning framework that combines curiosity driven sampling with a physics-informed quality control filter based on Simple Harmonic Oscillator model fits, allowing the system to automatically exclude low fidelity data during acquisition. Evaluations on a pre-acquired dataset of band-excitation piezoresponse spectroscopy (BEPS) data from PbTiO$_3$ thin films with spatially localized noise show that the proposed method outperforms random sampling, standard active learning, and multitask learning strategies. We further deployed the framework in real-time experiments on a separate PbTiO$_3$ thin-film sample with heterogeneous domain structures, demonstrating its effectiveness in autonomous microscopy experiments. Overall, this work supports hybrid autonomy in self-driving labs, where physics-informed quality assessment and active decision making work together for more reliable discovery.}




\maketitle

%
\section{Introduction}

\rev{Autonomous experimentation platforms are transforming the landscape of materials research by enabling rapid, closed-loop discovery cycles that combine physical instrumentation with machine learning (ML)~\cite{seifrid2022autonomous, macleod2020self, stach2021autonomous, vasudevan2021autonomous}. These self-driving labs promise to reduce the time and cost required to explore tasks such as learning structure--property relationships across large experimental design spaces. Determining these relationships effectively and accurately is vital to develop materials with desired properties~\cite{tom2024self, szymanski2023autonomous, epps2021accelerated, liu2022exploring}. However, their effectiveness is often constrained by a fundamental challenge-- the presence of low-quality, noisy data, which can mislead learning algorithms and degrade model performance. This problem is particularly pronounced in structure--property learning tasks such as Image-to-Spectrum (Im2Spec) and Spectrum-to-Image (Spec2Im) translations ~\cite{kalinin2021toward}, where both inputs and labels can be affected by spectral degradation arising from low signal-to-noise ratio (SNR), measurement instability, or localized experimental artifacts~\cite{borodinov2019deep, somnath2015full}.}

Active learning is an intelligent strategy that has been widely adopted in ML tasks where the goal is to reduce the number of exploration steps to learn a target function~\cite{ren2023autonomous, narasimha2024multiscale}. It can be applied to structure--property learning tasks like Image-to-Spectral or Spectral-to-Image predictions where the idea is to make careful decisions in the problem space to sample data that could be most beneficial to the learning model. Prior work explored this direction using deep encoder-decoder architectures and deep kernel learning to capture latent structure--property relationships and guide automated exploration~\cite{kalinin2021toward, roccapriore2021predictability, liu2022experimental}. Normally, the acquisition decision is made by estimating how much uncertainty a sample can reduce or how much information gain it offers. This approach can significantly reduce the number of samples needed, thereby saving resources and minimizing unnecessary sample degradation~\cite{settles2009active, ziatdinov2022bayesian, azimi2012batch}. In most cases, a standard approach to guide active learning is through a Gaussian Process (GP) predictor. The GP maps each sample to a level of uncertainty, and we aim to acquire samples where the predicted uncertainty is relatively high. This ensures that the model learns from samples that offer the most information, enabling learning over the full sample space with the fewer exploration steps~\cite{riis2022bayesian, sauer2023active, ziatdinov2022physics, zhao2020promoting}.

However, GP-based strategies have several limitations when used to determine optimal sample points for acquisition in specific scenarios. First, we often need a scalarizer to quantify the target of interest~\cite{pratiush2025scientific, hickman2025atlas}, which can be difficult to define before running an experiment~\cite{harris2025active, liu2023learning, liu2022experimental}. Second, while GP-based strategies are suitable for optimization, they often limit effective learning across high-dimensional spaces such as images or spectra. This limitation becomes particularly pronounced when only a few initial seed points are available, which is common in active learning scenarios~\cite{binois2022survey, altman2018curse}.

A recent study on curiosity-driven active learning~\cite{vatsavai2025curiosity} offers a practical solution to these problems. It uses a surrogate model to predict the margin of mistake of the base model (e.g., Im2Spec or Spec2Im), and samples from regions where this surrogate error model predicts higher error, effectively guiding exploration toward where the base model is likely to fail. While this approach works well under ideal conditions with little or no noise, it can fail in the presence of low-fidelity samples. Noisy data with very low signal-to-noise ratio (SNR) can be misinterpreted as highly informative by the surrogate model, since the base model struggles more in such regions and is likely to produce higher errors. This leads the learner to acquire misleading or uninformative samples, hindering the base model's learning process and performance. Prior work has shown that integrating physics-informed domain knowledge into the learning process can enhance the robustness and generalization capabilities of ML models, particularly under noisy or uncertain conditions~\cite{pratiush2025scientific, chowdhury2023evaluation, shen2023machine, chowdhury2025cglearn}. Recent novelty-oriented strategies such as INS2ANE~\cite{bulanadi2025beyond} also further emphasize the need for exploration that is not solely guided by uncertainty reduction. In this work, we propose a quality controlled acquisition strategy that leverages a physics-informed metric to gate low-fidelity samples during learning, enabling the model to focus on high-quality data and generalize more effectively in real-world, noise-prone environments. Conventional active learning methods typically ignore data fidelity and may mistakenly prioritize noisy measurements due to their high estimated uncertainty. Eventually, this not only harms the learning process but also leads to inefficient sampling~\cite{harris2025active}.

\rev{Building on this idea, we present \text{ActiveQC} (Active Learning with Quality Control), a gated active learning framework that integrates curiosity-driven acquisition with a quality control mechanism. ActiveQC uses a physics-informed quality metric to evaluate the fidelity of the acquired signal, which in this case is derived from Simple Harmonic Oscillator (SHO) fits across the Band Excitation Piezoresponse Spectroscopy (BEPS) spectra. This is used with Gaussian Process regression to filter out low-fidelity samples, ensuring that only high-quality data are added to the training pool. Notably, although Gaussian Process regression is used in this work, in principle, any surrogate model can be used in the gated active learning framework to facilitate the gating mechanism. 
We benchmark our approach on a pre-acquired paired dataset of piezoresponse force microscopy (PFM) images and BEPS data from ferroelectric PbTiO$_3$ thin films, with spatially localized spectral degradation introduced to simulate experimentally relevant low-fidelity measurement regions exhibiting spatial quality variations. Under this setting, the GP-based gating framework learns a spatial quality landscape to distinguish relatively low and high-fidelity regions during acquisition.}

\rev{Our evaluations span both Im2Spec and Spec2Im tasks, capturing different directions of the structure--property mapping. Through extensive experiments, we show that \text{ActiveQC} outperforms baseline strategies, including traditional random sampling, standard active learning, and multitask learning-based methods, by reducing error in the unseen space and improving reliability under data noise. We further qualitatively validated the framework through real-time Atomic Force Microscopy (AFM) experiments on a separate PbTiO$_3$ thin-film sample with richer domain structures and broader spectral-quality variations, demonstrating the practical feasibility of quality-controlled autonomous acquisition under experimental constraints. These results highlight the importance of incorporating data-quality awareness into acquisition pipelines and suggest a promising direction toward more reliable and physically meaningful autonomy in materials discovery.}

%
%
\section{Results}

We assess the effectiveness of the proposed ActiveQC framework on two reciprocal structure-property translation tasks: Image-to-Spectrum (Im2Spec) and Spectrum-to-Image (Spec2Im). These tasks mimic real-world experimental scenarios in autonomous microscopy, where localized measurement noise can significantly impact model generalization. 

\subsection{Data Preparation and Experimental Setup}

The pre-acquired dataset consists of co-registered PFM images and corresponding BEPS spectra sampled across a uniform spatial grid. Each training example includes a $16 \times 16$ image patch paired with a $256$-dimensional spectral vector extracted from the central coordinate of the corresponding image patch. The $256$-dimensional vector corresponds to the BEPS hysteresis loop, obtained by fitting the frequency-dependent complex BE response across multiple DC bias points. All inputs are normalized to the $[0,1]$ range to ensure numerical stability and consistency across training and evaluation. The full field of view consists of a $50 \times 50$ spatial grid. Using a stride of 1 pixel, we extract overlapping $16 \times 16$ patches, yielding a total of 1225 paired image-spectrum samples.

\begin{figure}[htbp]
\centering
\includegraphics[width=0.90\textwidth]{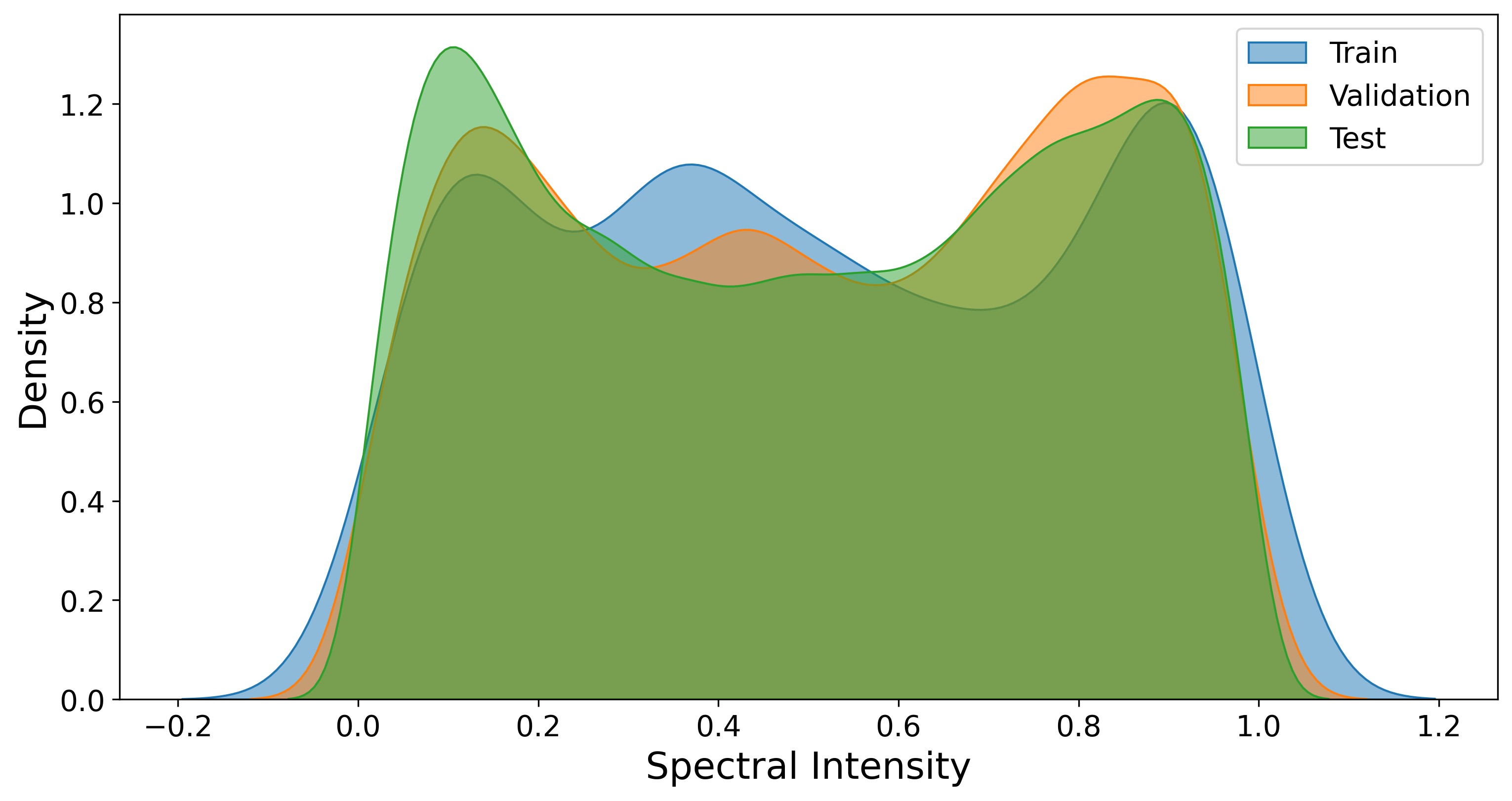}
\caption{Initial distribution of spectral intensities across the train, validation, and test subsets. Blue, orange, and green curves correspond to the train, validation, and test subsets, respectively. Kernel density estimates confirm that all splits preserve the overall spectral-intensity distribution, enabling fair and representative evaluation.}
\label{fig:res_gen_exp}
\end{figure}

%
To simulate a realistic low-resource setting reflective of practical experimental constraints, we adopt a sparse initialization: only 1\% of the dataset is used for initial training or as the seed samples, with an additional 9\% reserved for validation. The remaining 90\% serves as the unlabeled test pool.~\rev{For all quantitative evaluations, test errors are computed only on the clean (non-corrupted) test samples, while noisy/corrupted samples are excluded from evaluation. This allows the evaluation to specifically measure how effectively the models learn the underlying structure--property relationships rather than their ability to fit the synthetic degradation patterns.} At each acquisition iteration, 0.5\% of the pool is selected as a batch and added to the training set, allowing us to monitor how model performance evolves with increasing labeled data. As shown in Figure~\ref{fig:res_gen_exp}, the spectral intensity distributions across train, validation, and test sets remain well aligned, despite the small training fraction. This ensures that improvements are attributable to the acquisition strategy rather than dataset imbalance.

\begin{figure*}[htbp]
\centering
\includegraphics[width=0.98\textwidth]{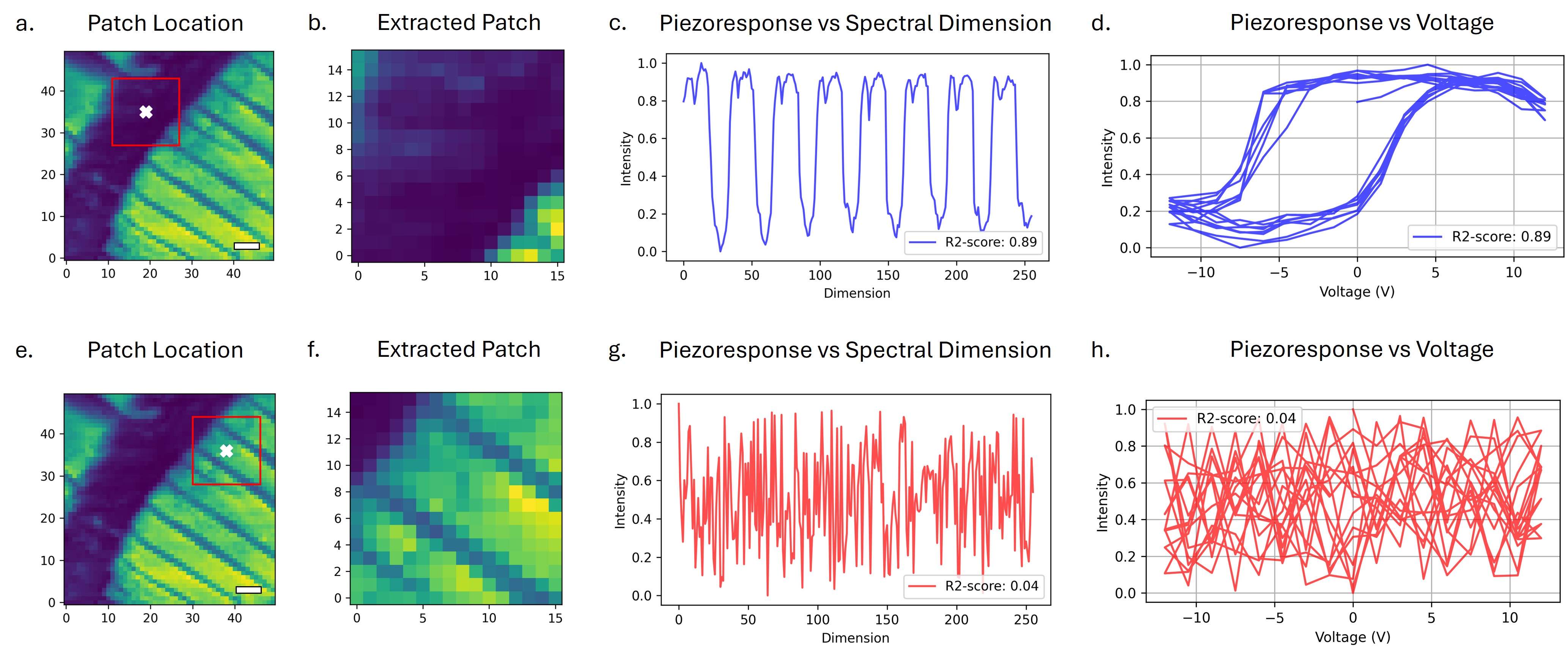}
\caption{Examples of two representative samples from clean and noise-impacted regions. Subplots (a)--(d) show the patch location, extracted image patch, piezoresponse spectrum over spectral dimension, and piezoresponse spectrum over applied voltage, respectively, for a sample unaffected by induced noise. Subplots (e)--(h) present the similar visualizations for a sample from a noise-affected region. The noisy spectra exhibit severe deviations from the expected ferroelectric hysteresis loop shape, leading to very low SHO-fit $R^2$-scores and highlighting the importance of incorporating quality-aware filters during learning and acquisition. Scale bar indicates a length of 100 nm.}
\label{fig:res_gen_pat}
\end{figure*}

\begin{figure*}[htbp]
\centering
\includegraphics[width=0.98\textwidth]{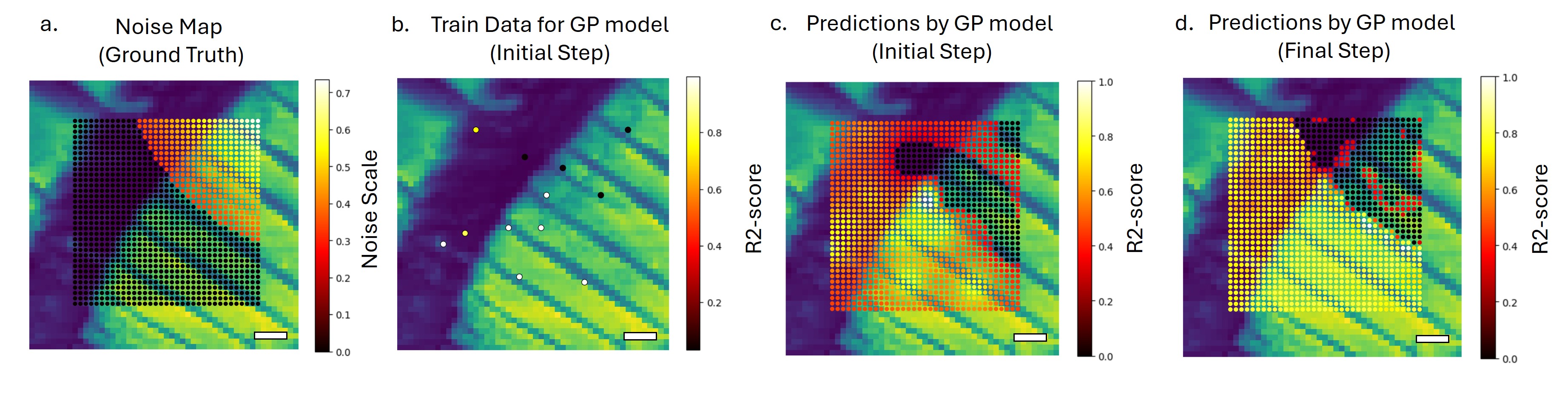}
\caption{Quality estimation using Gaussian Process regression.
(a) Ground-truth noise map applied to the BEPS spectra, where brighter colors indicate stronger corruption. (b) Mean SHO-fit $R^2$-scores computed for the initial set of sampled locations, based on fitting complex BEPS spectra across all DC-bias points. (c) GP-predicted spatial map of spectral quality (mean $R^2$) at the initial acquisition step. (d) GP-predicted quality map after the final acquisition step, showing improved delineation between high- and low-fidelity regions. Scale bar indicates a length of 100 nm.~\rev{Separate colorbars are shown for each panel for better visualization and interpretation as the displayed quantities correspond to different value ranges and distributions.}}
\label{fig:res_gen_noi}
\end{figure*}

\subsection{Simulating Spatially Localized Spectral Noise}

%
\rev{To simulate degradation observed in experimental microscopy pipelines, we introduce spatially structured Gaussian noise into the spectral domain. The simulated degradation specifically targets localized corruption of spectral fidelity through additive Gaussian noise and is intended to mimic low-SNR or degraded spectral measurements. Specifically, we target designated regions (e.g., the top-right quadrant), degrading spectral fidelity in a spatially localized manner. Approximately 30\% of the dataset is corrupted, creating a challenging benchmark for evaluating active acquisition strategies. The proposed GP-based quality gating framework assumes that spectral quality exhibits spatial correlation across the sample and consequently, the study focuses on spatially localized degradation patterns that are learned through the GP quality landscape.}

Figure~\ref{fig:res_gen_pat} shows representative examples from clean and corrupted regions. The clean sample retains well-defined spectral features and characteristic voltage hysteresis, while the noisy sample exhibits high-frequency artifacts and erratic voltage responses. This contrast highlights the severity of degradation that models must either robustly learn from or strategically avoid.

\subsection{Spectral Quality Estimation via SHO-$R^2$-GP Filtering}

To incorporate physics-informed robustness into the acquisition pipeline, we model the spatial distribution of spectral fidelity using a GP trained on quality scores derived from the SHO model~\cite{borodinov2019deep} fits. Each BEPS spectrum is decomposed into its frequency-dependent complex response at every DC-bias point, and an SHO model is independently fit to each frequency sweep. The mean $R^2$-score across all bias points serves as a physically grounded quality metric that reflects how well the measured response conforms to expected SHO dynamics. Additional details of the SHO fitting procedure and quality-metric construction are provided in the Supplementary Information (Section S1).

The GP is trained to regress these mean $R^2$-score values across spatial coordinates, enabling the system to estimate spectral reliability throughout the entire sample. During acquisition, candidates whose predicted SHO $R^2$-score falls below a predefined threshold ($\tau$) are gated out, preventing the active learner from selecting low-quality spectra.
 
%
As shown in Figure~\ref{fig:res_gen_noi}, the GP’s predictions become progressively more accurate as additional high-quality training data is collected. The initial estimate (Figure~\ref{fig:res_gen_noi}c) captures only coarse structures of the underlying noise distribution, whereas the final estimate (Figure~\ref{fig:res_gen_noi}d) sharply delineates regions of high and low spectral fidelity. Figure~\ref{fig:res_gen_noi}a presents the ground-truth noise field used to degrade approximately 30\% of the spectra, while Figure~\ref{fig:res_gen_noi}b shows the initial set of mean SHO $R^2$-score values used as training data. Overall, the learned quality landscape enables reliable filtering of low-fidelity data and ensures that subsequent acquisitions focus on high-quality, information-rich regions.

\begin{figure*}[htbp]
\centering
\includegraphics[width=0.98\textwidth]{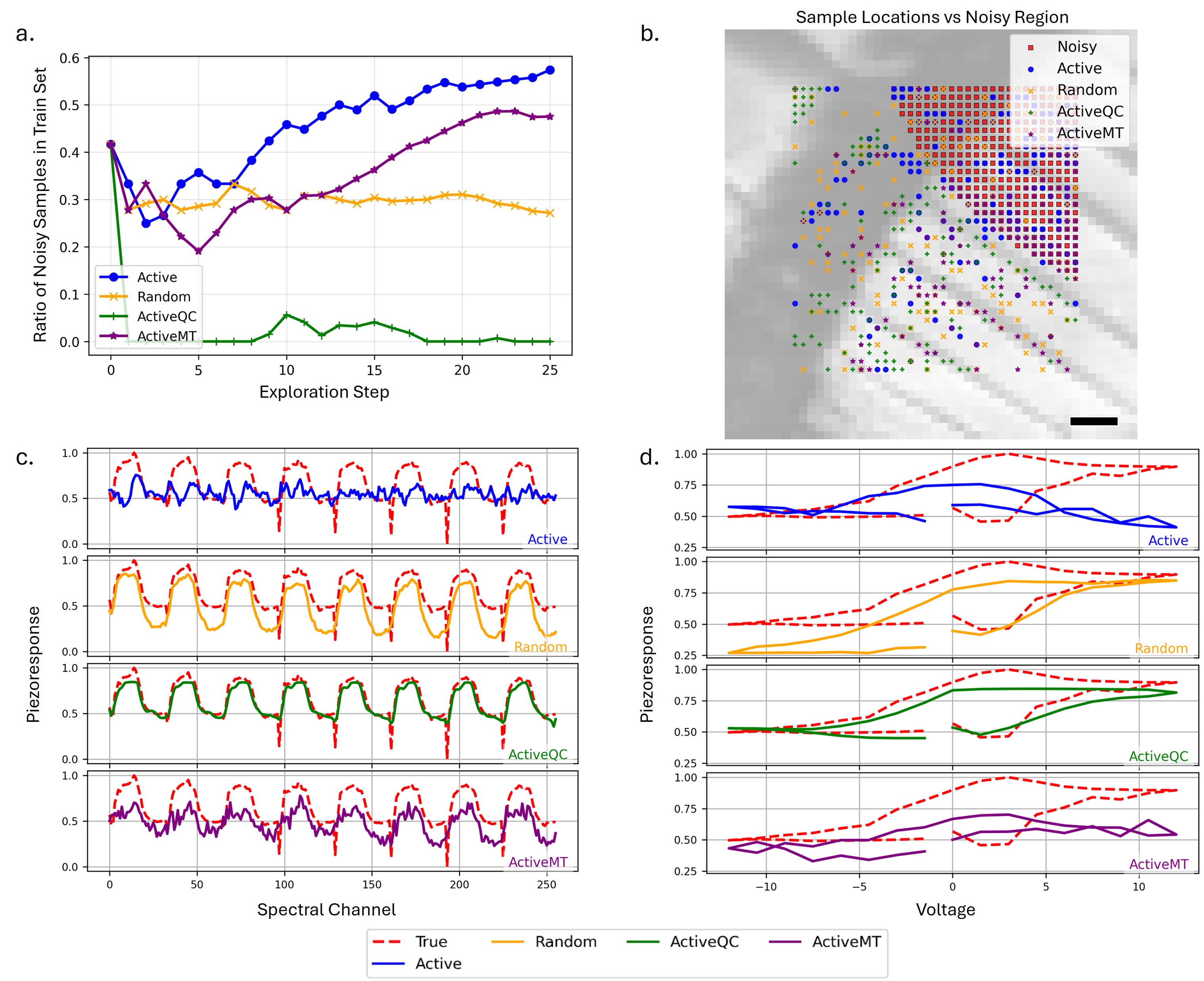}
\caption{Comparison on the Im2Spec task. (a) Ratio of noisy samples in the training set over exploration steps. Blue circles, orange crosses, green plus symbols, and purple stars correspond to Active, Random, ActiveQC, and ActiveMT strategies, respectively. (b) Sample locations overlaid on the PFM image after the final acquisition step. Red squares indicate the noisy region, while colored markers indicate sampled locations for each acquisition strategy. (c) Predicted piezoresponse plotted against spectral channel for a representative test sample. (d) The same predictions plotted against applied voltage. Red dashed curves indicate the ground-truth response, while blue, orange, green, and purple curves correspond to Active, Random, ActiveQC, and ActiveMT predictions, respectively. Scale bar indicates a length of 100 nm.}
\label{fig:res_i2s_nlp}
\end{figure*}

\begin{figure*}[htbp]
\centering
\includegraphics[width=0.98\textwidth]{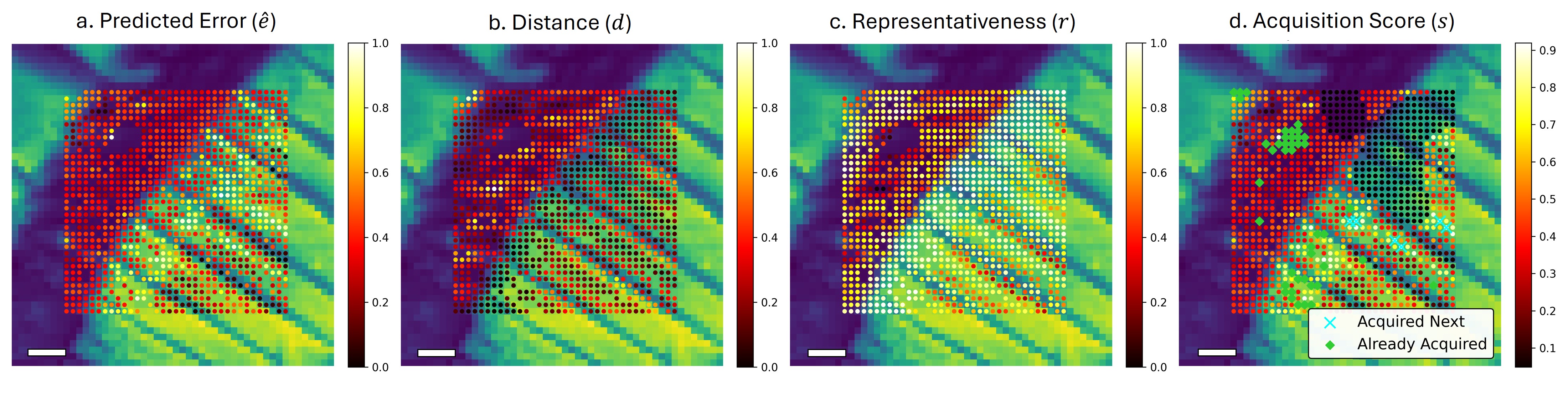}
\caption{Visualization of the components influencing ActiveQC sampling decisions for the Im2Spec task. (a) Predicted error by the error model for each test sample, reflecting regions where the Im2Spec model is most uncertain and could benefit from additional training data. (b) Distance from each test point to the current training set, encouraging exploration of underrepresented areas. (c) Representativeness of each test point with respect to the full unseen pool, favoring points that best summarize the remaining distribution. (d) Final acquisition score combining panels (a--c) using a weighted sum with coefficients $\alpha = 0.90$, $\beta = 0.05$, and $\gamma = 0.05$, together with gating applied via the quality predictor. Cyan crosses indicate samples selected for acquisition in the next exploration step, while green diamonds indicate already acquired samples. The gating mechanism suppresses samples in the noisy top-right region by assigning them very low acquisition scores, preventing their selection even when their predicted error, distance, or representativeness metrics are high. The visualization corresponds to an intermediate exploration step (Step 6). Scale bar indicates a length of 100 nm.}
\label{fig:res_i2s_eds}
\end{figure*}

\begin{figure}[htbp]
\centering
\includegraphics[width=0.80\textwidth]{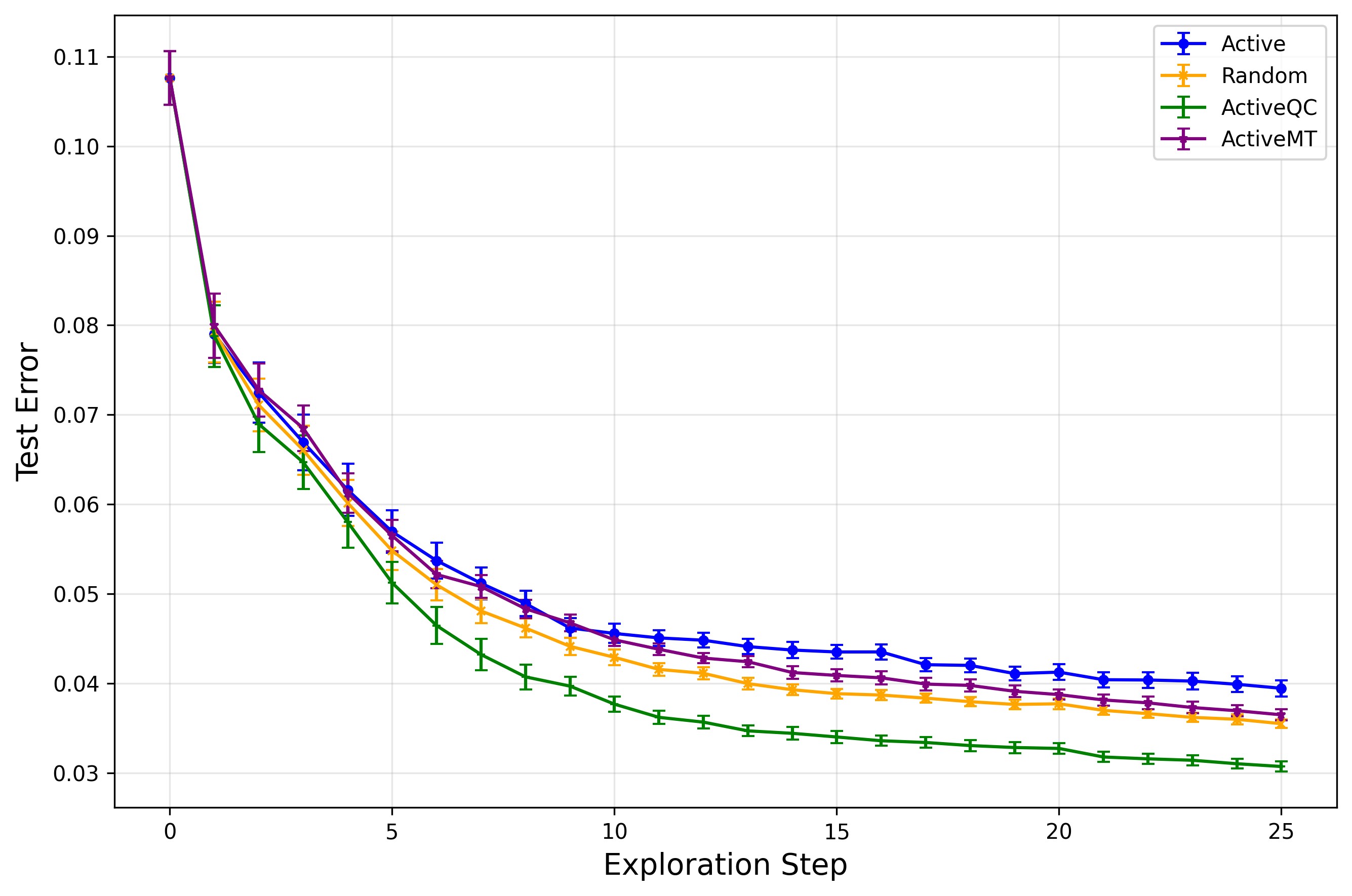}
\caption{Test error (mean squared error) over exploration steps on the Im2Spec task. Blue, orange, green, and purple curves correspond to Active, Random, ActiveQC, and ActiveMT strategies, respectively. Error bars represent the standard error of the mean (SEM) computed over 30 random trials.}
\label{fig:res_i2s_emu}
\end{figure}

\subsection{Im2Spec: Image-to-Spectrum Translation}

In this section, we discuss our results on the Image-to-Spectrum translation task. To understand the sampling behavior of different strategies, Figure~\ref{fig:res_i2s_nlp}(a) shows the ratio of noisy samples acquired during training across exploration steps by different strategies. Both Active and ActiveMT learning strategies progressively select more noisy samples, as the base Im2Spec model tends to interpret these challenging regions as highly informative. Interestingly, ActiveMT-with an auxiliary Image-to-Image task partially mitigates this bias, but the improvement is limited. Random maintains a relatively stable noise ratio due to its unbiased nature. In contrast, ActiveQC starts with a few noisy acquisitions but quickly learns to avoid them via its SHO-$R^2$-based quality gating. Figure~\ref{fig:res_i2s_nlp}(b) further illustrates the spatial acquisition patterns overlaid on the PFM image background. ActiveQC distinctly avoids the top-right region, which was the noisy quadrant, whereas other strategies-particularly Active and ActiveMT sample heavily from this area. Figures~\ref{fig:res_i2s_nlp}(c) and~\ref{fig:res_i2s_nlp}(d) visualize predictions on a representative test sample from each strategy. While Active, Random, and ActiveMT show the impact of having noisy samples in the training pool in their predictions, ActiveQC produces a cleaner reconstruction closely matching the true signal. This highlights how quality-aware acquisition leads to better learning of structure-property relationships. \rev{For all Im2Spec benchmark experiments presented in this section, the quality threshold was fixed at $\tau = 0.30$ throughout all acquisition steps and trials.}

Figure~\ref{fig:res_i2s_eds} offers an interpretable breakdown of the acquisition logic behind ActiveQC at an intermediate exploration step during the active learning process. Subplot~\ref{fig:res_i2s_eds}(a) shows the predicted error for each unseen sample. This guides the exploitation component by highlighting uncertain regions that the base Im2Spec model struggles with. Subplot~\ref{fig:res_i2s_eds}(b) visualizes the total distance from each candidate test point to the existing training set, enabling exploration of spatially diverse and less familiar areas. Subplot~\ref{fig:res_i2s_eds}(c) illustrates the representativeness of each sample, defined as the similarity of a candidate test point to other candidate test points. This promotes coverage of the unseen distribution by prioritizing samples that are central to the remaining pool. Finally, subplot~\ref{fig:res_i2s_eds}(d) presents the computed acquisition score that combines the three metrics via a weighted sum using Eq.~\ref{eqn:met_acq_scr}, with gating applied by the SHO-$R^2$-based quality control applied through the gating rule in Eq.~\ref{eqn:met_acq_fin}.

Figure~\ref{fig:res_i2s_emu} presents the mean squared error (MSE) across acquisition steps for all the different strategies.~\rev{As seen in this figure, under the spatially localized degradation scenarios considered here, ActiveQC consistently outperforms the baselines across the acquisition trajectory.} In early rounds, Active and ActiveMT briefly reduce test error, but their performance quickly plateaus or degrades due to the inclusion of low-quality (noisy) samples. In contrast, ActiveQC effectively filters out such degraded spectra via a learned SHO-$R^2$-based quality gate, allowing the acquisition budget to focus exclusively on high-fidelity regions. 

\begin{table}[htbp]
\small
\caption{\ Mean and standard error of the mean (SEM) of test errors at the last three acquisition rounds on the Im2Spec task, averaged across 30 random trials. ActiveQC significantly outperforms all baselines ($p < 0.05$, Welch's t-test).}
\label{tbl:res_i2s_emu}
\begin{tabular}{lccc}
  \hline
  \textbf{Method} & \textbf{Step 23} & \textbf{Step 24} & \textbf{Step 25} \\
  \hline
  Random   & 0.0362 ± 0.0005 & 0.0360 ± 0.0006 & 0.0355 ± 0.0005 \\
  Active   & 0.0402 ± 0.0009 & 0.0399 ± 0.0009 & 0.0394 ± 0.0009 \\
  ActiveMT & 0.0373 ± 0.0006 & 0.0369 ± 0.0006 & 0.0365 ± 0.0006 \\
  ActiveQC & \textbf{0.0314 ± 0.0006} & \textbf{0.0310 ± 0.0005} & \textbf{0.0307 ± 0.0006} \\
  \hline
\end{tabular}
\end{table}

Statistical analysis over the final three acquisition rounds confirms that the performance improvements of ActiveQC over Random, Active, and ActiveMT are statistically significant (Welch’s t-test, $p < 0.05$), as illustrated in Table~\ref{tbl:res_i2s_emu}. This validates that quality-controlled sampling is not only intuitively beneficial, but also yields quantifiable gains in model reliability.

~\rev{We additionally evaluated the sensitivity of ActiveQC to the choice of the quality threshold $\tau$ across multiple values ranging from permissive to highly restrictive gating regimes. Detailed analysis for the Im2Spec task is provided in Supplementary Information (Section S2).}

\subsection{Spec2Im: Spectrum-to-Image Translation}

\begin{figure*}[htbp]
\centering
\includegraphics[width=0.98\textwidth]{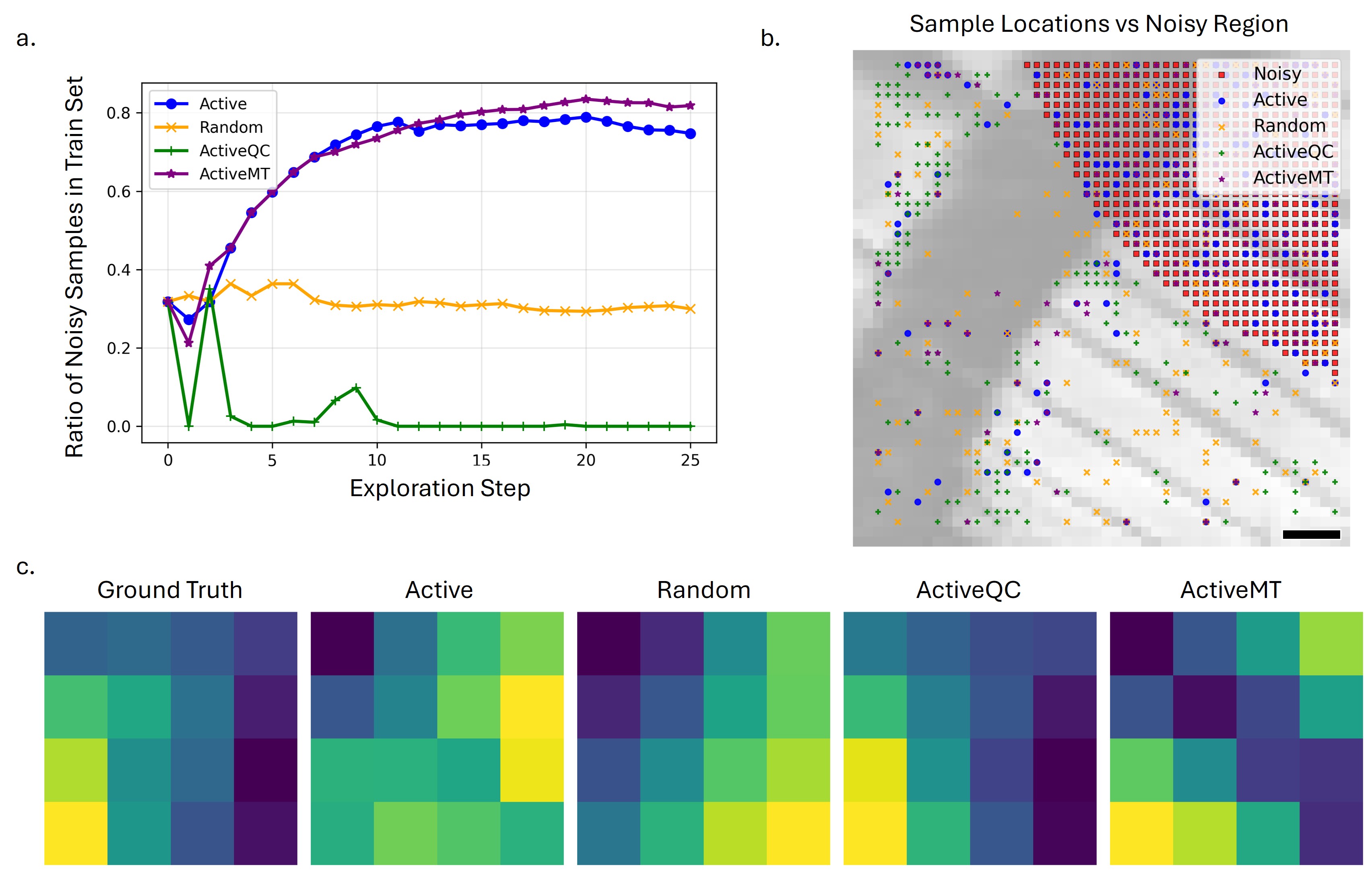}
\caption{Comparison on the Spec2Im task. (a) Ratio of noisy samples in the training set over exploration steps. Blue circles, orange crosses, green plus symbols, and purple stars correspond to Active, Random, ActiveQC, and ActiveMT strategies, respectively. (b) Spatial acquisition patterns overlaid on the PFM image after the final acquisition step. Red squares indicate the noisy region, while colored markers indicate sampled locations for each acquisition strategy. (c) Predicted image patches for a representative test sample. From left to right, panels correspond to the ground-truth image patch, followed by predictions from Active, Random, ActiveQC, and ActiveMT strategies, respectively. Scale bar indicates a length of 100 nm.}
\label{fig:res_s2i_nlp}
\end{figure*}

\begin{figure*}[htbp]
\centering
\includegraphics[width=0.98\textwidth]{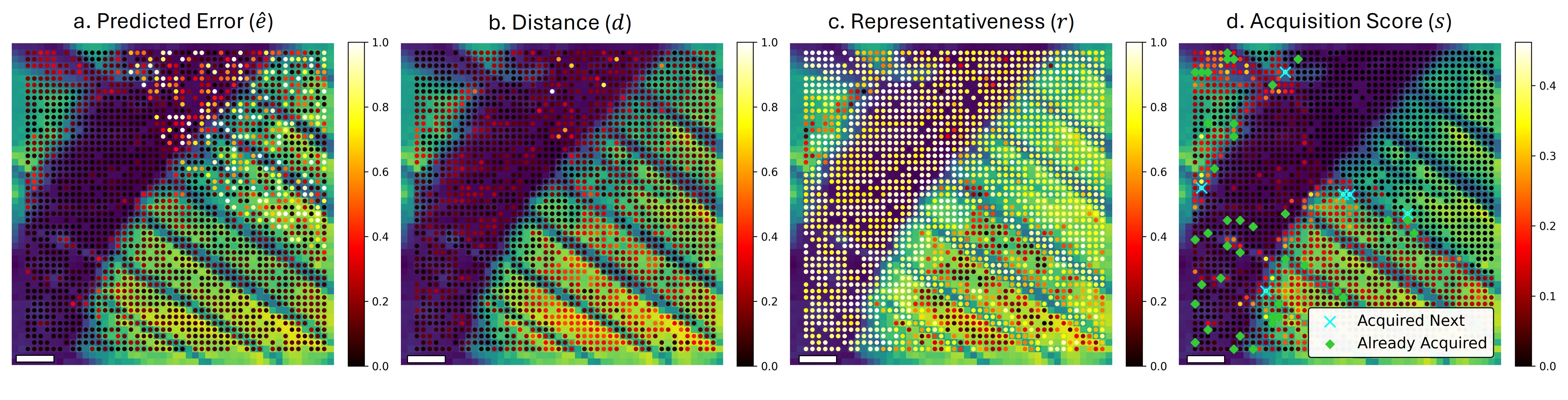}
\caption{Visualization of the components influencing ActiveQC sampling decisions in the Spec2Im task. (a) Predicted error by the error model for each candidate test sample, indicating regions where spectral inputs yield the most uncertain image predictions. (b) Distance from each test spectrum to the current training pool, encouraging exploration beyond already observed regimes. (c) Representativeness of each spectrum with respect to the full set of unseen spectra. (d) Final acquisition score combining panels (a--c) using a weighted sum with coefficients $\alpha = 0.90$, $\beta = 0.05$, and $\gamma = 0.05$, together with gating applied via the quality predictor. Cyan crosses indicate samples selected for acquisition in the next exploration step, while green diamonds indicate already acquired samples. The visualization corresponds to an intermediate exploration step (Step 6). Similar to the Im2Spec case, the gating mechanism down-weights the noisy top-right region by assigning minimal acquisition scores and preventing those samples from being selected. Separate colorbars with different intervals are intentionally shown for each panel for better visualization and interpretation, since the displayed quantities correspond to different value ranges and distributions, particularly panel (d), which represents the weighted acquisition score. Scale bar indicates a length of 100 nm.}
\label{fig:res_s2i_eds}
\end{figure*}

\begin{figure}[htbp]
\centering
\includegraphics[width=0.85\textwidth]{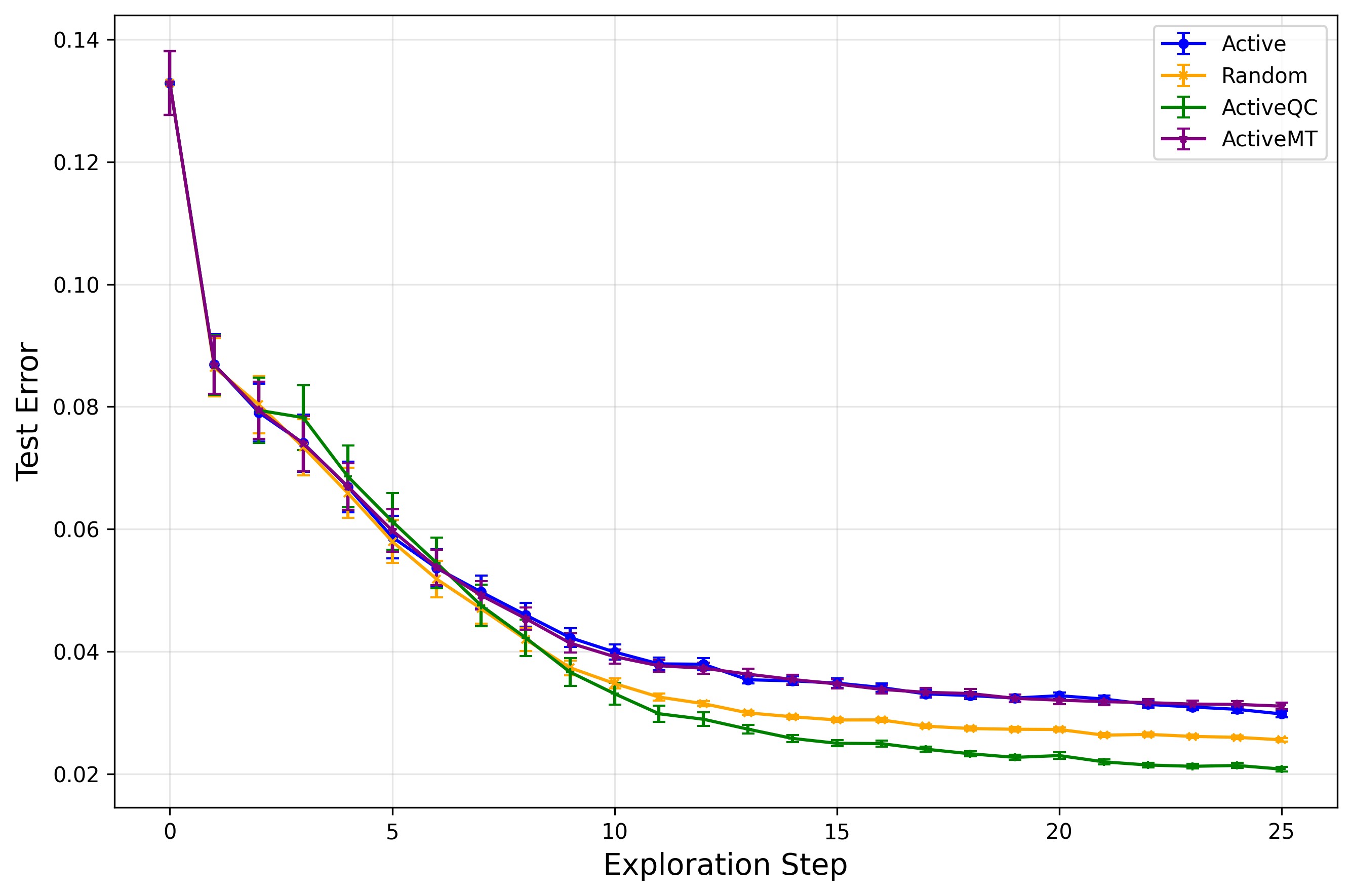}
\caption{Test mean squared error over exploration steps on the Spec2Im task. Blue, orange, green, and purple curves correspond to Active, Random, ActiveQC, and ActiveMT strategies, respectively. Error bars represent the standard error of the mean (SEM) computed over 30 random trials. ActiveQC maintains the lowest error throughout, especially during the later exploration rounds.}
\label{fig:res_s2i_emu}
\end{figure}

To further analyze acquisition behavior, we followed similar experiments with the inverse task (Spectrum-to-Image translation), where the goal is to map each 256-dimensional BEPS spectrum to a smaller $4\times4$ PFM image patch. This reduced output resolution reflects the higher difficulty of the inverse mapping and stabilizes training, since predicting full $16\times16$ patches directly from spectra is significantly more challenging regardless of the acquisition strategy used. Figure~\ref{fig:res_s2i_nlp}(a) shows that ActiveMT relying on a corrupted auxiliary spectral decoder accumulates the highest fraction of noisy samples. Random sampling exhibits moderate bias, while ActiveQC rapidly suppresses degraded inputs through its SHO-$R^2$-based gating mechanism. Figure~\ref{fig:res_s2i_nlp}(b) visualizes the spatial acquisition behavior. Similar to the Im2Spec experiments, we see ActiveQC distinctly avoids the upper-right quadrant, which contains high-noise spectra, while other strategies fail to discriminate this region. The qualitative impact of this filtering is evident in the reconstructed PFM patches (Figure~\ref{fig:res_s2i_nlp}(c)). ActiveQC recovers more cleaner, high-fidelity images that closely resemble the ground truth. However, compared to Im2Spec, the impact of noise in Spec2Im is noticeably milder. This is because noise affects only the inputs (spectra) rather than the targets, and the encoder-decoder architecture partially suppresses input noise during feature extraction.~\rev{Similar to the previous experiment, for all Spec2Im benchmark experiments presented in this section, the quality threshold was fixed at $\tau = 0.30$ throughout all acquisition steps and trials.}

%

%
The acquisition decomposition in Figure~\ref{fig:res_s2i_eds} affirms that ActiveQC balances model-driven exploration with robust quality filtering. Candidates that exhibit high curiosity (uncertainty/error), diversity (distance), or representativeness of the unexplored pool but has low SHO-$R^2$ quality scores are avoided during training and acquisition. This interplay enables the model to remain exploratory while ensuring that only high-fidelity spectra contribute to learning.

Figure~\ref{fig:res_s2i_emu} tracks test MSE over time. While all methods initially reduce error, only ActiveQC maintains consistent improvement in later stages-where noisy samples become more prominent in the learning task. This demonstrates its ability to extract meaningful structure-property relationships while suppressing noise-induced variability.

\begin{table}[htbp]
\small
\caption{\ Mean and standard error of the mean of the test error at the final three acquisition rounds on the Spec2Im task, averaged across 30 random trials. ActiveQC consistently achieves significantly lower error than all baseline strategies ($p < 0.05$, Welch's t-test).}
\label{tbl:res_s2i_emu}
\begin{tabular}{lccc}
  \hline
  \textbf{Method} & \textbf{Step 23} & \textbf{Step 24} & \textbf{Step 25} \\
  \hline
  Random   & 0.0261 ± 0.0003 & 0.0260 ± 0.0003 & 0.0256 ± 0.0003 \\
  Active   & 0.0309 ± 0.0005 & 0.0305 ± 0.0005 & 0.0298 ± 0.0006 \\
  ActiveMT & 0.0314 ± 0.0006 & 0.0313 ± 0.0005 & 0.0311 ± 0.0005 \\
  ActiveQC & \textbf{0.0212 ± 0.0004} & \textbf{0.0214 ± 0.0004} & \textbf{0.0208 ± 0.0004} \\
  \hline
\end{tabular}
\end{table}

Finally, Table~\ref{tbl:res_s2i_emu} confirms that the margin of improvement is statistically significant with ActiveQC compared to all other baselines, reinforcing the advantage of incorporating quality-awareness directly into the selection loop.

\rev{A corresponding sensitivity analysis was also performed for the Spec2Im task across multiple $\tau$ values and detailed results are provided in Supplementary Information (Section S2).}

\subsection{Real-time AFM Deployment}

\rev{Building on the results obtained from pre-acquired datasets, we next deployed the proposed quality-controlled active learning framework on an operating AFM for real-time adaptive sampling. To evaluate robustness under realistic experimental conditions, we selected a separate PbTiO$_3$ (PTO) thin-film sample with heterogeneous ferroelectric domain structures. An epitaxial (100)/(001)-oriented PbTiO$_3$ (PTO) thin film (45~nm thick) with a mixed a/c-domain structure was grown on a (100)$(\mathrm{Ba}_{0.25}\mathrm{Sr}_{0.75})\mathrm{RuO}_3$ // (100)$\mathrm{KTaO}_3$ substrate using pulsed metal-organic chemical vapor deposition. Detailed growth conditions are described in a previous study~\cite{ehara2022film}. All experiments were performed using the AEcroscopy platform~\cite{liu2024aecroscopy}, enabling closed-loop control of data acquisition, model updates, and decision making.}

\begin{figure*}[htbp]
\centering
\includegraphics[width=0.98\textwidth]{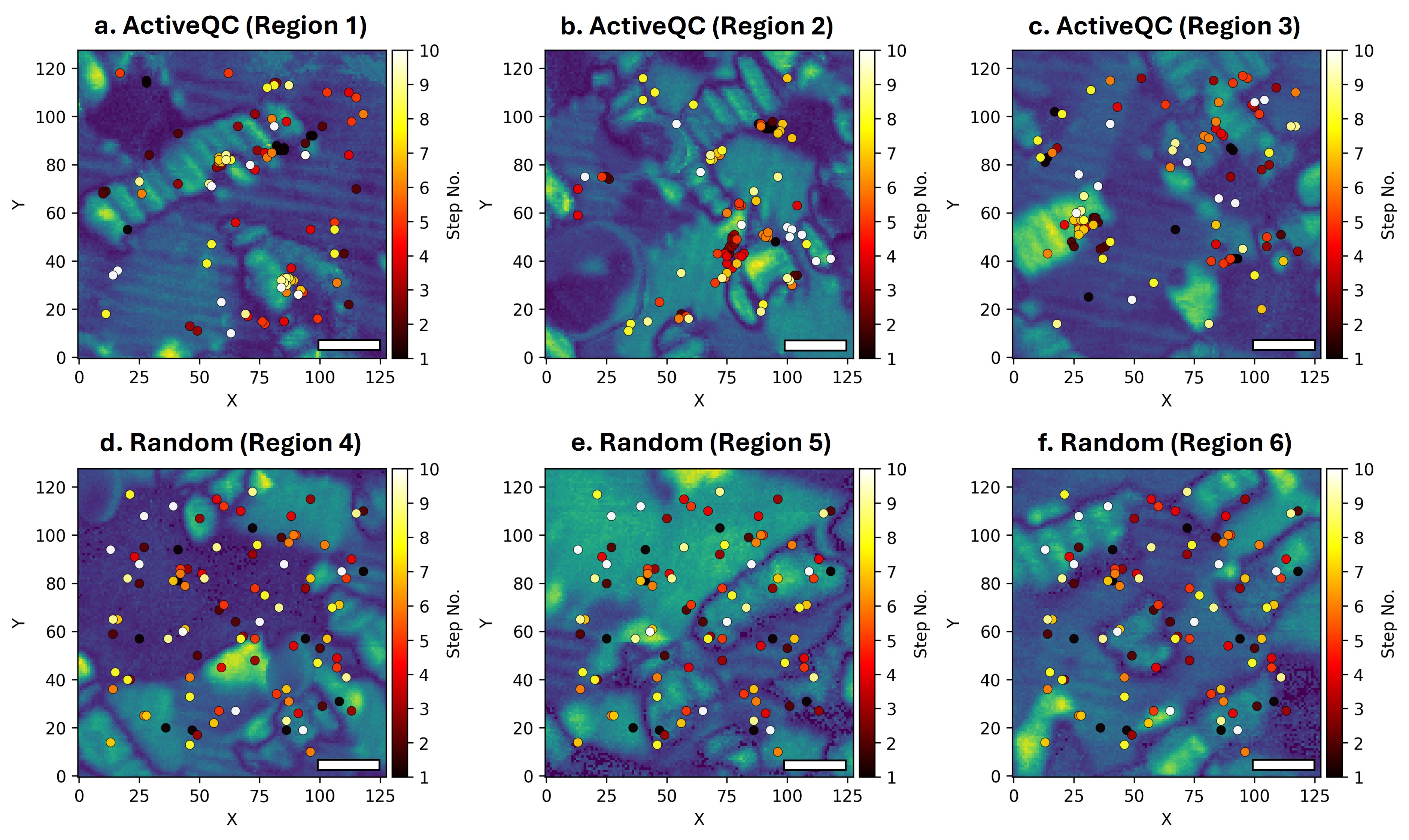}
\caption{\rev{Real-time AFM deployment of ActiveQC and Random sampling strategies across multiple PTO regions. Panels~(a--c) show three independent regions explored using the proposed ActiveQC strategy, while panels~(d--f) show three independent regions explored using Random sampling. Colored markers indicate the spatial acquisition sequence across exploration steps. In the ActiveQC cases, sampling locations become concentrated mostly near structurally active regions e.g. near domain walls. In contrast, Random sampling distributes acquisitions uniformly across the field of view due to the random nature of acquisition. Scale bars correspond to 400~nm.}}
\label{fig:res_rti_afm_1}
\end{figure*}

\rev{Unlike the pre-acquired benchmark experiments, real-time AFM deployment does not provide access to the full ground-truth information in advance. Consequently, quantitative comparisons such as test-set MSE cannot be computed during acquisition. Evaluation in the real-time setting therefore relies primarily on qualitative assessment guided by domain expertise and physical interpretation of the acquired ferroelectric responses. Furthermore, due to irreversible or partially irreversible switching dynamics induced by repeated electrical biasing in ferroelectric measurements, identical regions cannot be repeatedly sampled using different acquisition strategies for direct one-to-one comparison because the applied electric fields can modify the local ferroelectric state through switching events, charge injection, vacancy redistribution, electrochemical reactions, and other partially irreversible processes~\cite{vasudevan2017ferroelectric}. In other words, the system lacks complete return point memory~\cite{sethna_return_point_memory}. Similar practical limitations have also been discussed in prior autonomous ferroelectric microscopy studies~\cite{bulanadi2025beyond} and therefore, to minimize region-specific bias, multiple independent regions were explored separately using different sampling strategies.}

\rev{The exploration trajectories for the six sampled regions are summarized in Fig.~\ref{fig:res_rti_afm_1}. Panels~(a--c) correspond to three independent PTO regions explored using the proposed ActiveQC framework, while panels~(d--f) show another three independent regions explored using Random sampling. Colorbar indicates the exploration step during acquisition. In the ActiveQC cases, we found the acquisition strategy adaptively concentrates measurements around spatially informative regions, particularly near domain walls, where polarization transitions between neighboring ferroelectric domains give rise to more complex electromechanical behavior. In contrast, Random sampling distributes acquisitions more uniformly across the field of view without considering spectral quality or acquisition utility due to its random nature.} 

%
\begin{figure*}[htbp]
\centering
\includegraphics[width=0.98\textwidth]{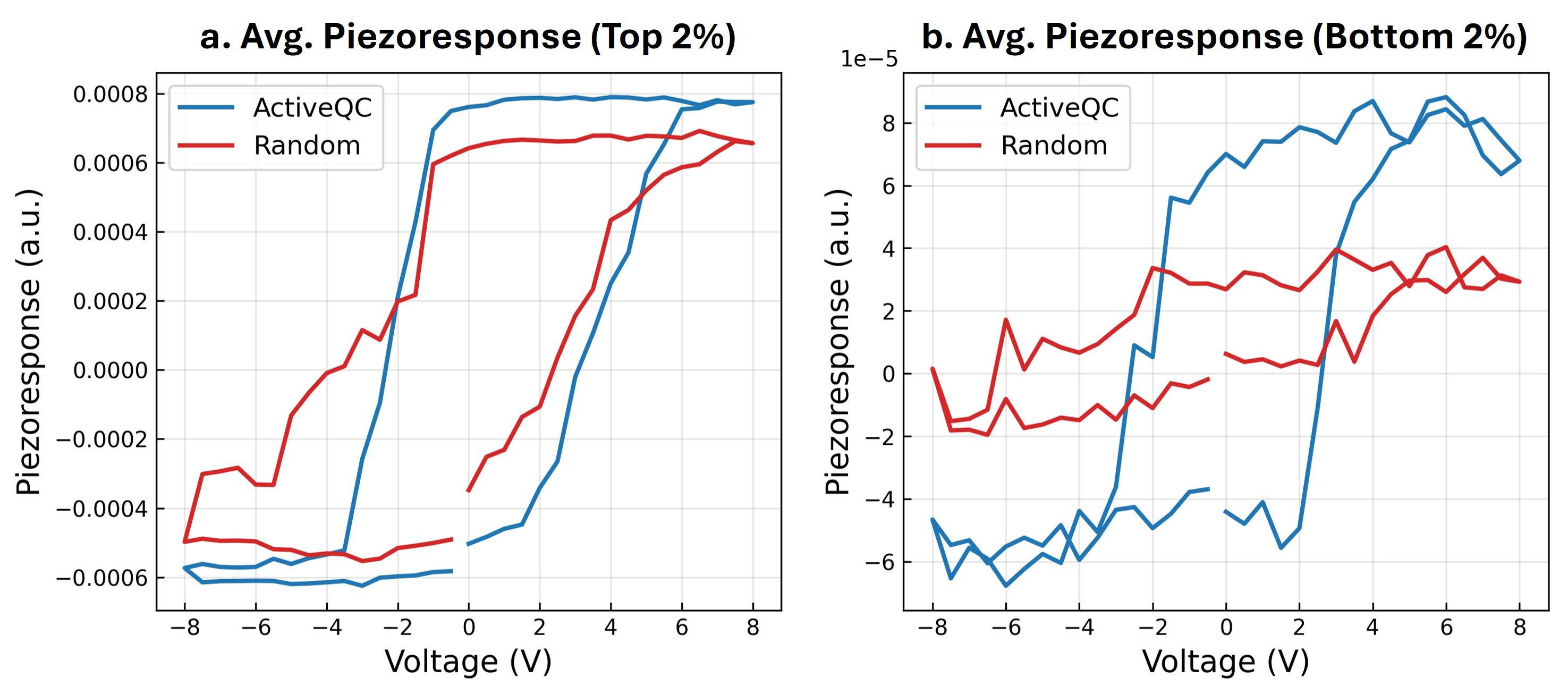}
\caption{\rev{Qualitative comparison of averaged piezoresponse hysteresis loops acquired using ActiveQC and Random sampling during real-time AFM deployment. Blue curves correspond to ActiveQC and red curves correspond to Random sampling. (a) Averaged piezoresponse loops computed from the top $2\%$ spectra with the largest hysteresis loop areas acquired during the final five exploration steps, where ranking was performed separately within each acquisition strategy. The ActiveQC strategy produces spectra with stronger ferroelectric switching characteristics. (b) Corresponding averaged piezoresponse loops computed from the bottom $2\%$ spectra with the smallest hysteresis loop areas acquired during the final five exploration steps. Compared to Random sampling, ActiveQC qualitatively demonstrates an improved ability to select loops that appear to show stronger switching characteristics and avoid regions that show weak switching.}}
\label{fig:res_rti_afm_2}
\end{figure*}

\rev{To further evaluate the impact of quality-controlled acquisition on the measured ferroelectric responses, Fig.~\ref{fig:res_rti_afm_2} compares the averaged piezoresponse hysteresis behavior obtained using ActiveQC and Random sampling. Since no full ground-truth dataset is available during real-time deployment, the comparison is performed qualitatively using physically meaningful ferroelectric switching characteristics evaluated by domain expertise. Panel~(a) shows the averaged piezoresponse hysteresis loops computed from the top $2\%$ spectra with the largest hysteresis loop areas acquired during the final five exploration steps, where the spectra were ranked separately within each acquisition strategy. Blue curves correspond to the averaged top $2\%$ spectra acquired using ActiveQC, while red curves correspond to the averaged top $2\%$ spectra acquired using Random sampling. As shown in Fig.~\ref{fig:res_rti_afm_2}(a), both strategies are capable of acquiring physically meaningful spectra; however, the ActiveQC strategy produces responses exhibiting stronger ferroelectric switching characteristics.}

%
\rev{Panel~(b) shows the corresponding averaged piezoresponse responses computed from the bottom $2\%$ spectra with the smallest hysteresis loop areas acquired during the final five exploration steps, again ranked separately within each acquisition strategy. Blue curves correspond to the averaged bottom $2\%$ spectra acquired using ActiveQC, while red curves correspond to the averaged bottom $2\%$ spectra acquired using Random sampling. In this regime, the distinction between the two acquisition strategies becomes substantially more pronounced. Spectra acquired using Random sampling exhibit significantly degraded hysteresis behavior with substantially suppressed piezoresponse signals. In contrast, although lower loop area samples are still present within the ActiveQC acquisitions, the corresponding responses remain more physically meaningful and less degraded overall. These results qualitatively demonstrate that the proposed quality-control mechanism can suppress acquisition in locations where the measured switching is weak and reduce the likelihood of low-fidelity spectra contaminating downstream tasks such as learning structure--property relations. During real-time deployment, the quality threshold ($\tau$), corresponding to the SHO-$R^2$ score used for acquisition gating, was selected empirically by the domain expert at each exploration stage based on the observed spectral responses, associated quality scores, and the desired level of quality enforcement. Detailed visualizations of the individual exploration steps, GP-predicted quality maps, acquisition-value maps, sampled locations, and quality thresholds used during the real-time experiments are provided in Supplementary Information (Section~S3).}

%
~\rev{Although quantitative evaluation is not possible in the absence of ground-truth measurements during live deployment, qualitative assessment based on domain expertise indicates that integrating quality control with curiosity-driven acquisition improves acquisition reliability and reduces the prevalence of severely degraded spectra during autonomous experimentation.}
%
%
\section{Discussion}

In this work, we introduced a quality-controlled active learning framework for structure--property prediction tasks in autonomous materials characterization. Our method combines curiosity-driven acquisition with a gating mechanism based on physics-informed quality metrics, specifically, the mean $R^2$ score from Simple Harmonic Oscillator (SHO) model fits and Gaussian process regression to exclude low-fidelity samples during training and acquisition. By leveraging these physics-based quality filters, the proposed method prevents corrupted inputs or labels from degrading model performance, a failure mode that traditional active learning approaches often fall into and even exacerbate over time.

We evaluated the proposed strategy on two bidirectional prediction tasks involving paired structural and spectral data. In both directions, our method consistently outperformed random sampling, standard active learning, and multitask learning baselines across multiple trials. Statistical tests confirmed that these gains were significant, particularly in later acquisition rounds when prediction models stabilize and the detrimental effects of low-quality data become more pronounced. Our analysis further showed that the impact of noise differs depending on whether it affects the inputs or the labels. While a multitask learner can partially mitigate these effects by leveraging auxiliary tasks, its improvements remain limited, and a dedicated gating mechanism is essential for maintaining a robust prediction pipeline. We also explored alternative acquisition metrics such as distance from the training set and representativeness of the unlabeled pool, to balance exploration and exploitation. These components ensure that the learner prioritizes curiosity, diversity, coverage, and data fidelity throughout the acquisition process. Finally, we validated the framework in real-time AFM experiments on a separate PbTiO$_3$ thin film with heterogeneous domain structures, demonstrating its robustness under practical conditions and its potential for seamless integration into autonomous experimental workflows.~\rev{However, the present study specifically focuses on localized spectral degradation and low-fidelity spectral measurements, and additional work will be required to investigate other experimental artifacts such as positional drift/distortion, or temporally correlated measurement degradation.}

~\rev{Overall, we found that integrating quality assessment into active learning pipelines can enhance acquisition reliability. In the present work, the quality-control mechanism is specifically based on SHO-model fitting and GP-based quality estimation, which are physically meaningful for BEPS measurements and band-excitation scanning probe microscopy. We therefore want to emphasize that the current experimental validation is limited to BEPS-based ferroelectric characterization tasks. However, the broader concept of incorporating quality-aware gating into active learning pipelines may still be transferable to other scientific domains where suitable physics-informed, signal-based, or domain-specific quality metrics can be defined. Examples may include reference-based matching, signal-to-noise thresholds, learned quality estimators, or other physically interpretable quality measures tailored to the underlying experimental modality. More broadly, this work underscores the value of hybrid autonomy, where physics-informed and/or domain knowledge based priors complement algorithmic exploration to accelerate and de-risk discovery in autonomous research systems.}
%
\section{Methods}

The relationships between material structure and ferroelectric properties explored in our study are investigated using band excitation piezoresponse force microscopy (BE-PFM) measurements on PbTiO$_3$ thin films. Structural information is derived from BE-PFM amplitude and phase maps, while corresponding functional properties are obtained from BEPS signals. The preacquired dataset and measurement procedures closely resemble those used in prior work~\cite{vatsavai2025curiosity}.~\rev{Additional details regarding BEPS measurements, PFM imaging, and SHO-based fitting procedures can be found in prior works~\cite{liu2022experimental, borodinov2019deep}.} We assess our method through two bidirectional tasks namely Im2Spec and Spec2Im, leveraging spatially resolved measurements across the sample.

\subsection{Dataset Overview}

Our experiments utilize a dataset comprising paired structural and spectroscopic measurements from PbTiO$_3$ thin films. Each sample includes a local $16\times16$ PFM image patch and its corresponding BEPS spectrum, enabling both forward and inverse structure--property mappings. The patch size is selected based on prior studies~\cite{vatsavai2025curiosity} to reflect the spatial extent over which local structure influences spectral behavior.

To simulate realistic experimental degradation, we induce Gaussian noise into spatially localized regions of the dataset, selectively corrupting specific areas. In these regions, the spectral data are intentionally degraded. For the Im2Spec task, spectral targets are corrupted and for Spec2Im, spectral inputs are corrupted while target image patches remain clean. This decoupling allows us to independently assess the model sensitivity to input versus output data degradation.

\subsection{Acquisition Strategies}

We evaluated four acquisition strategies under a pool-based active learning setting, where an initial training seed is iteratively expanded with new samples drawn from the unexplored pool:

\begin{itemize}
    \item \textbf{Random:} Serves as a control baseline in which new acquisitions are selected uniformly at random, without considering model uncertainty or data quality.

    \item \textbf{Active:} A curiosity-driven acquisition strategy previously proposed~\cite{vatsavai2025curiosity}. It trains a surrogate model to estimate the error made by the base model (e.g., Im2Spec or Spec2Im) using latent features extracted from the encoder. This strategy captures model uncertainty and guides acquisition toward informative samples in ideal, low-noise scenarios.

    \item \textbf{ActiveMT:} A multi-task (MT) variant that incorporates auxiliary input reconstruction. An additional decoder branch reconstructs the input to regularize the latent space. For Im2Spec, this involves reconstructing the image patch; for Spec2Im, the spectrum. This setup discourages overfitting to noisy labels by promoting consistency in the latent space to serve multiple tasks.

    \item \textbf{ActiveQC:} Our proposed quality-controlled (QC) gated sampling strategy. ActiveQC integrates curiosity-driven acquisition with a physics-informed quality control mechanism based on the Simple Harmonic Oscillator model fits. For each candidate BEPS spectrum, the SHO model is fit to all DC bias points, and a quality score is computed as the mean coefficient of determination ($R^{2}$) over the individual frequency-domain fits. A Gaussian Process (GP) is trained on these $R^{2}$ scores to model the spatial distribution of data fidelity. During acquisition, samples whose predicted $R^{2}$ falls below a predefined threshold are excluded from being selected. By gating out low-fidelity spectra, ActiveQC prevents corrupted inputs or labels from misleading the surrogate error model and ensures that only high-quality data contribute to model training.
\end{itemize}

In our experiments, all strategies were initialized with the same seed set and gathered samples in batches at each iteration according to their respective criteria.

\begin{figure*}[htbp]
    \centering
    \includegraphics[width=0.98\textwidth]{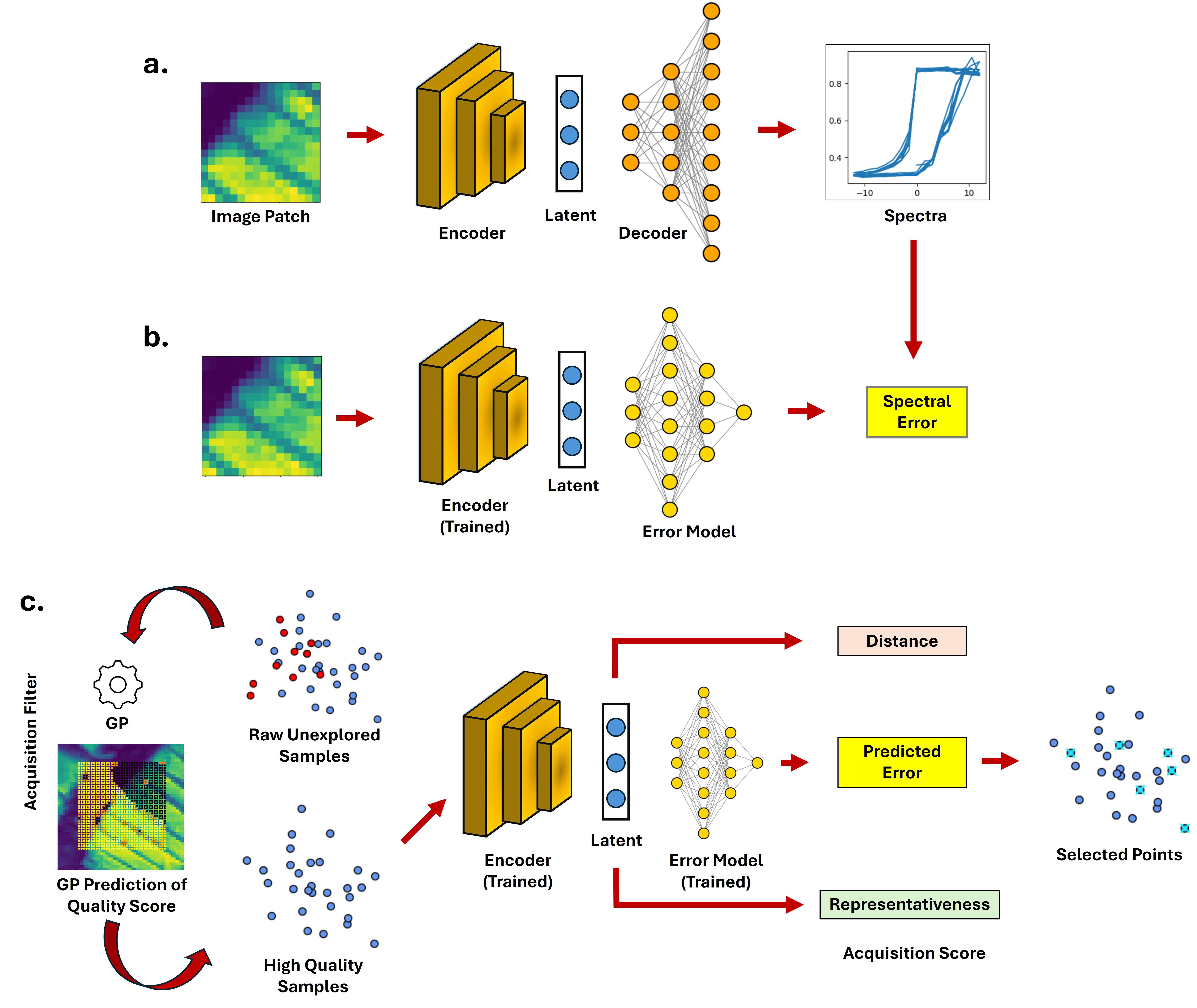}
    \caption{Overview of the proposed ActiveQC framework for the Im2Spec task. 
    (a) Training of the base Im2Spec model: an input image patch is passed through an encoder-decoder network to generate the corresponding spectrum, from which the spectral prediction error is computed. 
    (b) Training of the surrogate error model: using the latent representation produced by the trained encoder, a separate model is trained to predict the spectral error estimated in panel (a). 
    (c) ActiveQC acquisition pipeline: a Gaussian Process (GP)-based quality filter predicts spectral fidelity across the spatial domain and removes low-quality candidates from the acquisition pool. For the remaining high-quality samples, an acquisition score is computed by combining predicted error, distance to the existing training set, and representativeness of each candidate. Top-ranked candidates are then selected for exploration in the next iteration.}     
    \label{fig:met_i2s_aqc}
\end{figure*}

\begin{figure*}[htbp]
    \centering
    \includegraphics[width=0.98\textwidth]{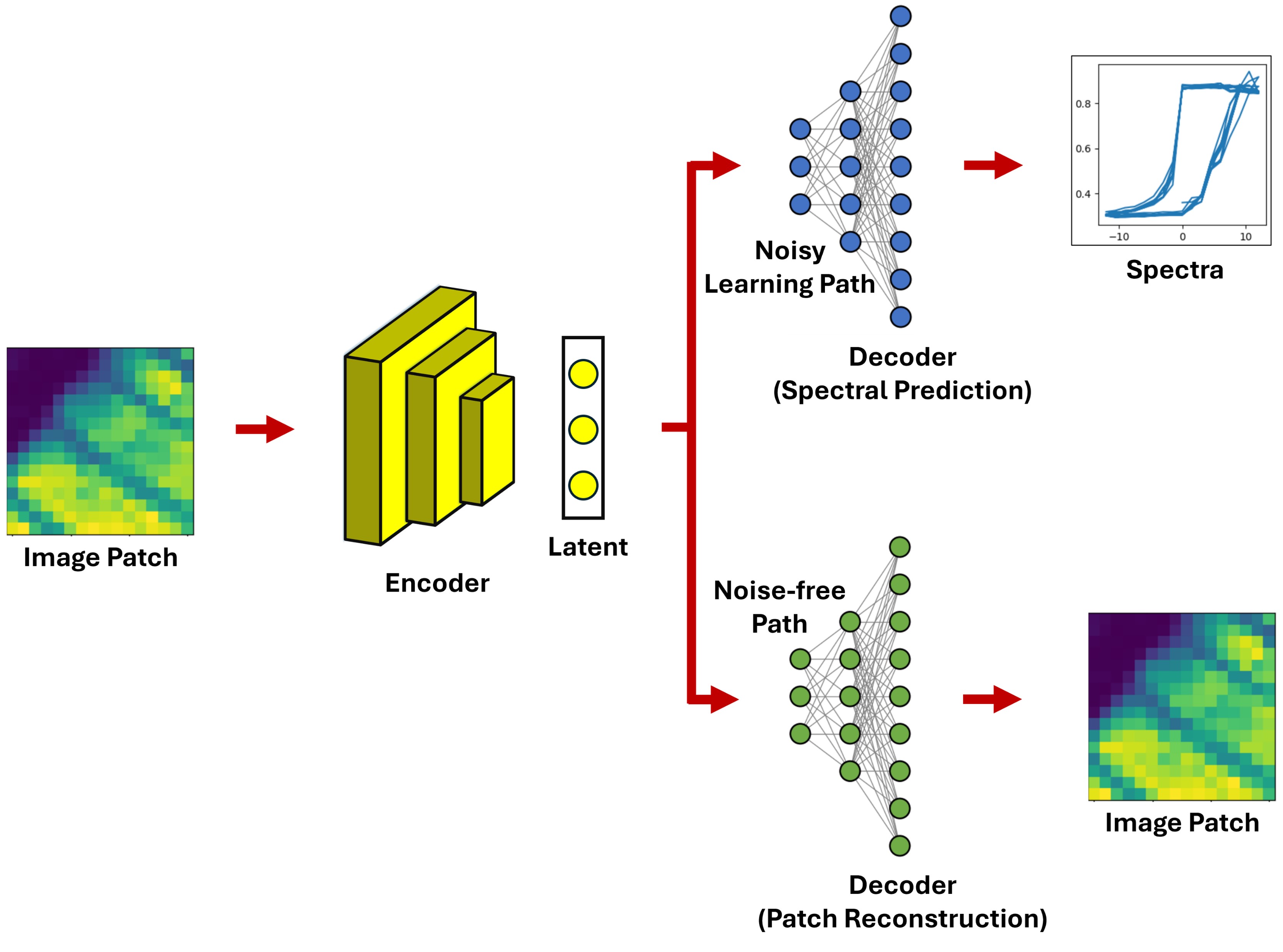}
    \caption{Overview of the ActiveMT (multitask learner) baseline framework for the Im2Spec task. An input image patch is encoded into a latent representation, which is passed through two decoder branches. The upper blue branch predicts spectra and is trained using noisy spectral measurements. The lower green branch reconstructs the original input image, providing a clean supervision signal to regularize the latent space. This multitask design mitigates the impact of noise and promotes more robust learning.}    
    \label{fig:met_i2s_amt}
\end{figure*}

\subsection{Model Architectures}

This section details the model designs underlying the ActiveQC and ActiveMT strategies. 
Both approaches are built upon an encoder--decoder architecture tailored for structure--property mapping in the Im2Spec and Spec2Im tasks.

Figure~\ref{fig:met_i2s_aqc} illustrates our proposed ActiveQC pipeline for the Im2Spec task. 
The encoder transforms a $16\times16$ PFM image patch into a compact latent representation, which the decoder uses to predict the corresponding BEPS spectrum.  A surrogate error model is trained to estimate the base model’s spectral error using these latent embeddings. To ensure that only high-fidelity spectra are acquired, ActiveQC incorporates a physics-informed gating mechanism. For every candidate BEPS spectrum, the SHO model is fit to each DC bias point, and a mean $R^{2}$ score is computed across all frequency-domain fits. A GP is trained to model the spatial distribution of these quality scores, enabling prediction of data fidelity across the sample. 
During acquisition, samples whose predicted $R^{2}$ falls below a predefined threshold are filtered out, preventing noisy or corrupted spectra from entering the training pipeline. Acquisition scores for the remaining candidates are computed by combining the surrogate error prediction, spatial distance to the current training set, and representativeness within the unseen pool. The highest-scoring candidates are selected, sampled and added to the training set in each iteration.

Figure~\ref{fig:met_i2s_amt} shows the ActiveMT architecture for the Im2Spec task. 
It extends the base encoder-decoder model with a multitask decoder: the main decoder predicts spectra (which may be noisy), while an auxiliary decoder reconstructs the input PFM image patch, which remains uncorrupted. This dual-path training regularizes the latent space by anchoring it with a clean auxiliary task, improving robustness to label noise in the spectral domain and enhancing generalization under data corruption. During active learning, the latent representations produced by this multitask-trained encoder are used by the surrogate error model to estimate the spectral prediction error across candidate samples, guiding the selection of new acquisition points.

\subsection{Acquisition Criteria: Balancing Exploration vs Exploitation}

Once low-quality samples are gated out using the physics-informed GP quality model, three acquisition factors are computed for each remaining candidate sample:

\begin{itemize}
    \item \textbf{Predicted Error:} Estimated by the surrogate model, capturing the base model’s expected error.
    \item \textbf{Distance:} Euclidean distance in the latent space to the current training set, encouraging exploration of underrepresented regions.
    \item \textbf{Representativeness:} Cosine similarity to the remaining unseen pool, promoting coverage of the unexplored space.~\rev{Specifically, representativeness is computed in the encoder latent space by summing the cosine similarities between the latent embedding of a candidate sample and the latent embeddings of the remaining unseen pool. Samples with higher similarity scores are considered more representative of the unexplored distribution and can therefore be prioritized to improve coverage of the unseen space.}
\end{itemize}

These are combined into a unified acquisition score:

\begin{equation}\label{eqn:met_acq_scr}
    s_i = \alpha \cdot \hat{e}_i + \beta \cdot d_i + \gamma \cdot r_i,
\end{equation}

where $\hat{e}_i$ is the predicted error, $d_i$ is the distance to the current training set, and $r_i$ measures representativeness among the unseen samples. \rev{The coefficients $\alpha$, $\beta$, and $\gamma$ are user-defined hyperparameters that control the balance between exploitation and exploration during acquisition. In this work, a larger weight was assigned to the predicted error term ($\alpha = 0.90$) to prioritize regions where the structure--property prediction model is uncertain, while smaller weights on distance and representativeness ($\beta = 0.05$, $\gamma = 0.05$) encourage some spatial diversity and coverage of the unseen pool. We kept these coefficients fixed throughout all the experimental steps.}

To ensure robustness in noisy experimental settings, ActiveQC incorporates a gating criterion based on the mean $R^{2}$ value computed from the SHO fits performed at every bias point in the BEPS spectrum, where each $R^{2}$ score quantifies how well the complex frequency-domain BE response is explained by the SHO model. A Gaussian Process (GP) is trained to model the spatial distribution of spectral quality scores and predict the fidelity of each candidate spectrum. The final acquisition rule is thus:

\begin{equation}\label{eqn:met_acq_fin}
a_i =
\begin{cases}
    s_i & \text{if } q_i \ge \tau, \\
    0   & \text{otherwise},
\end{cases}
\end{equation}

where \rev{$q_i$ is the GP-predicted SHO-$R^{2}$-based quality score of candidate $i$, and $\tau$ is a predefined SHO-$R^{2}$ quality threshold selected empirically by the domain expert at each exploration stage based on the observed spectral responses, associated quality scores, and the desired level of quality enforcement to suppress low-fidelity measurements.} Samples predicted to have low spectral fidelity are therefore assigned negligible acquisition priority, preventing them from being selected even if they appear highly uncertain or representative. This gating mechanism ensures that acquisitions remain both informative and physically meaningful, leading to more stable and robust structure-property learning.

%

%
\section*{Data Availability}

Upon acceptance, the datasets generated and analyzed during the current study will be made publicly available on Zenodo: \url{https://doi.org/10.5281/zenodo.18881435}.

\section*{Code Availability}

Upon acceptance, code and notebooks for data preparation, model training, active learning experiments, and analysis will be made publicly available at: \url{https://github.com/hasanjawad001/curiosity-driven-gated-active-learning}.

The structure-property learning ML models used in this work utilize the Im2Spec framework~\cite{kalinin2021toward} together with its associated open-source repository (\url{https://github.com/ziatdinovmax/im2spec}), as well as the AtomAI open-source repository (\url{https://github.com/pycroscopy/atomai}). The deployment of the workflows on the AFM was interfaced using AEcroscopy (\url{https://github.com/yongtaoliu/aecroscopy.pyae}).

\section*{Acknowledgements}
Algorithmic development for curiosity-driven gated active learning was supported by the US Department of Energy, Office of Science, Office of Basic Energy Sciences, MLExchange Project, award number 107514. Microscopy research was supported by the Center for Nanophase Materials Sciences (CNMS), which is a US Department of Energy, Office of Science User Facility at Oak Ridge National Laboratory, and conducted as part of a user project. This work (H.F.) was also supported by Japan Science and Technology Agency (JST) as part of Adopting Sustainable Partnerships for Innovative Research Ecosystem (ASPIRE) (No. JPMJAP2312), MEXT Initiative to Establish Next-generation Novel Integrated Circuits Centers (X-NICS) (No. JPJ011438) and by MEXT Program: Data Creation and Utilization Type, Material Research and Development Project (No. JPMXP1122683430).

\section*{Author Contributions}

\textbf{J.C.}: Investigation (lead); Software (lead); Formal Analysis (lead); Writing-Original Draft (lead). 
\textbf{G.N.}: Investigation (supporting); Software (supporting); Writing-Review \& Editing (supporting). 
\textbf{J.Y.}: Resources (supporting); Investigation (supporting); Writing-Review \& Editing (supporting). 
\textbf{H.F.}: Resources (supporting); Investigation (supporting); Writing-Review \& Editing (supporting). 
\textbf{Y.E.}: Resources (supporting); Investigation (supporting); Writing-Review \& Editing (supporting). 
\textbf{Y.L.}: Investigation (supporting); Software (supporting); Writing-Review \& Editing (supporting). 
\textbf{R.V.}: Conceptualization (lead); Supervision (lead); Investigation (lead, AFM experiments); Formal Analysis (supporting); Writing-Review \& Editing (lead). 
All authors read and approved the final manuscript.


\section*{Competing Interests}

R.V. is an Associate Editor of \textit{npj Computational Materials} and was not involved in the journal’s review of, or decisions related to, this manuscript. The other authors declare no competing financial or non-financial interests.

\end{document}


\maketitle

\section*{S1. SHO Fitting and $R^2$-Based Quality Metric}

Band-excitation piezoresponse spectroscopy (BEPS) data were analyzed using a physics-informed Simple Harmonic Oscillator (SHO) model to quantify the physical fidelity of the measured electromechanical response. At each spatial location, the complex frequency-dependent piezoresponse signal was recorded at multiple DC bias points along the hysteresis loop. An independent SHO fit was performed for every frequency sweep corresponding to each DC bias point.

The complex SHO response was modeled as
\begin{equation}
H(\omega) = A \, e^{i\phi} \, \frac{\omega_0^2}{\omega^2 - i \omega \omega_0 / Q - \omega_0^2},
\end{equation}
where $A$ is the response amplitude, $\omega_0$ is the resonance frequency, $Q$ is the quality factor, and $\phi$ is the phase offset. This formulation captures both the amplitude and phase behavior of the cantilever--sample interaction and is commonly used to describe BE-PFM dynamics~\cite{borodinov2019deep}.

To enable robust fitting of the complex-valued response, nonlinear least-squares optimization was used to estimate the SHO parameters. Physically motivated bounds and heuristic initial guesses ensured stable convergence. Fits were performed independently for each spatial pixel, excitation cycle, DC bias point, and field direction.

The quality of each SHO fit was quantified using the coefficient of determination ($R^2$),
\begin{equation}
R^2 = 1 - \frac{\sum_i \left( y_{\mathrm{obs},i} - y_{\mathrm{fit},i} \right)^2}
{\sum_i \left( y_{\mathrm{obs},i} - \bar{y}_{\mathrm{obs}} \right)^2},
\end{equation}
where $y_{\mathrm{obs}}$ and $y_{\mathrm{fit}}$ denote the observed and fitted responses (including both real and imaginary components), and $\bar{y}_{\mathrm{obs}}$ is the mean of the observed signal.

For each spatial location, an $R^2$ value was computed independently at every DC bias point along the hysteresis loop. The final spectral quality score was defined as the mean $R^2$ averaged across all DC bias points,
\begin{equation}
q = \frac{1}{N_{\mathrm{DC}}} \sum_{j=1}^{N_{\mathrm{DC}}} R^2_j.
\end{equation}
This scalar quality metric reflects the overall consistency of the measured spectrum with the SHO model and penalizes spectra affected by noise, distortion, or non-physical behavior. These mean SHO-$R^2$ scores were subsequently used as physics-informed quality labels for Gaussian process regression in our quality-controlled active learning framework.

\section*{\rev{S2. Sensitivity Analysis of the Quality Threshold $\tau$}}

\begin{figure*}[htbp]
    \centering
    \includegraphics[width=0.9\textwidth]{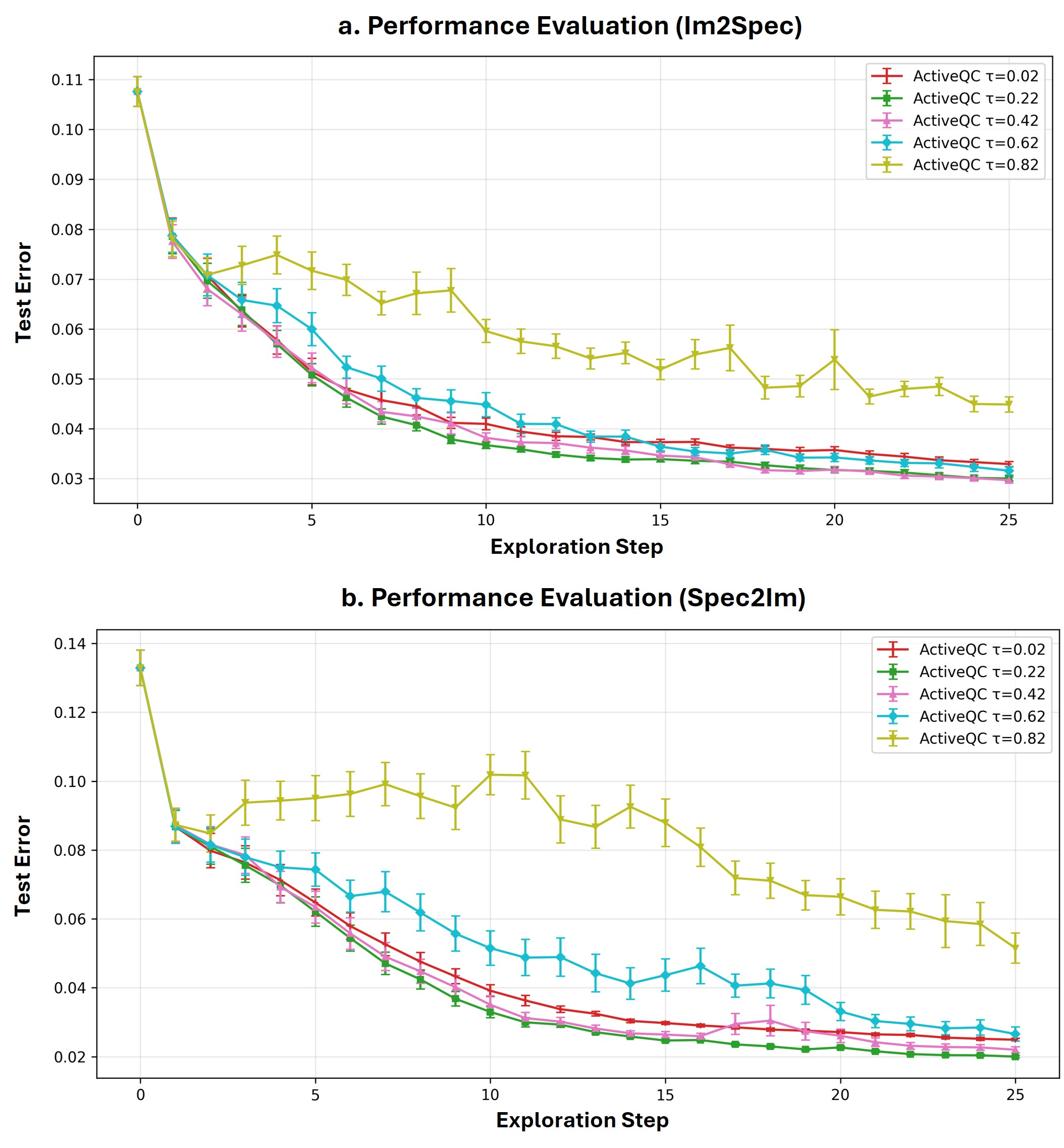}
    \caption{\rev{Sensitivity of ActiveQC performance to the quality threshold $\tau$. (a) Im2Spec task and (b) Spec2Im task across different threshold values, showing mean test error across 30 trials, with error bars representing the standard error of the mean (SEM). Red curves correspond to $\tau = 0.02$, green curves to $\tau = 0.22$, pink curves to $\tau = 0.42$, cyan curves to $\tau = 0.62$, and yellow curves to $\tau = 0.82$. Intermediate threshold values ($\tau = 0.22$ and $0.42$) consistently provide the strongest overall performance by balancing quality enforcement with sufficient acquisition diversity. In contrast, very permissive and highly restrictive thresholds lead to degraded or sub-optimal performance. These results further emphasize the importance of domain-expert guided threshold selection for quality control to balance suppression of low-fidelity measurements with preservation of useful exploratory diversity.}}    
    \label{fig_tau_sensitivity}
\end{figure*}

\begin{table*}[htbp]
\centering
\small
\caption{\rev{Sensitivity of ActiveQC performance to the quality threshold $\tau$. Mean $\pm$ standard error of the mean (SEM) of the test mean squared error (MSE) at the final three acquisition rounds for the Im2Spec and Spec2Im tasks, averaged across 30 trials. Intermediate threshold values ($\tau = 0.22$ and $0.42$) consistently provide the strongest overall performance across both tasks.}}
\label{tab:tau_sensitivity}
\begin{tabular}{ccccc}
\hline
\textbf{Task} & \textbf{Method} & \textbf{Step 23} & \textbf{Step 24} & \textbf{Step 25} \\
\hline
Im2Spec & ActiveQC $\tau=0.02$ & 0.0333 $\pm$ 0.0005 & 0.0329 $\pm$ 0.0005 & 0.0328 $\pm$ 0.0005 \\
Im2Spec & ActiveQC $\tau=0.22$ & \textbf{0.0301 $\pm$ 0.0005} & \textbf{0.0300 $\pm$ 0.0005} & \textbf{0.0297 $\pm$ 0.0004} \\
Im2Spec & ActiveQC $\tau=0.42$ & \textbf{0.0301 $\pm$ 0.0005} & \textbf{0.0297 $\pm$ 0.0006} & \textbf{0.0292 $\pm$ 0.0005} \\
Im2Spec & ActiveQC $\tau=0.62$ & 0.0323 $\pm$ 0.0009 & 0.0316 $\pm$ 0.0008 & 0.0318 $\pm$ 0.0014 \\
Im2Spec & ActiveQC $\tau=0.82$ & 0.0450 $\pm$ 0.0015 & 0.0449 $\pm$ 0.0015 & 0.0473 $\pm$ 0.0031 \\
\hline
Spec2Im & ActiveQC $\tau=0.02$ & 0.0252 $\pm$ 0.0005 & 0.0249 $\pm$ 0.0004 & 0.0249 $\pm$ 0.0004 \\
Spec2Im & ActiveQC $\tau=0.22$ & \textbf{0.0204 $\pm$ 0.0003} & \textbf{0.0200 $\pm$ 0.0003} & \textbf{0.0196 $\pm$ 0.0003} \\
Spec2Im & ActiveQC $\tau=0.42$ & 0.0227 $\pm$ 0.0008 & 0.0220 $\pm$ 0.0009 & 0.0214 $\pm$ 0.0008 \\
Spec2Im & ActiveQC $\tau=0.62$ & 0.0284 $\pm$ 0.0022 & 0.0265 $\pm$ 0.0021 & 0.0259 $\pm$ 0.0021 \\
Spec2Im & ActiveQC $\tau=0.82$ & 0.0585 $\pm$ 0.0062 & 0.0515 $\pm$ 0.0043 & 0.0506 $\pm$ 0.0046 \\
\hline
\end{tabular}
\end{table*}


\rev{To evaluate the sensitivity of ActiveQC to the choice of the quality threshold $\tau$, we repeated the simulated benchmark experiments using five threshold values: $\tau = 0.02$, $0.22$, $0.42$, $0.62$, and $0.82$. These values span a range from highly permissive gating, where most candidate samples pass the quality gate except the severly degraded ones, to highly restrictive gating, where many candidate samples are suppressed. Threshold values larger than $0.82$ were not considered because they filtered out most candidate samples, leaving very limited acquisition diversity and making the acquisition process nearly deterministic.}

\rev{Figure~\ref{fig_tau_sensitivity} summarizes the sensitivity analysis for both the Im2Spec and Spec2Im tasks. In both cases, intermediate thresholds ($\tau = 0.22$ and $\tau = 0.42$) provide the strongest overall performance, suggesting a favorable balance between suppressing low-fidelity spectra and preserving sufficient acquisition diversity. In contrast, both highly permissive and moderately restrictive thresholds ($\tau = 0.02$ and $\tau = 0.62$) lead to suboptimal performance for opposite reasons: permissive gating provides limited quality control, whereas restrictive gating begins to suppress otherwise useful candidate samples and reduce acquisition diversity. At $\tau = 0.82$, the threshold becomes overly restrictive, excessively suppressing useful high-quality candidate samples and leading to degraded and less stable performance across trials. These results further emphasize the importance of keeping the domain expert in the loop when selecting the quality threshold. A sufficiently permissive threshold may allow low-fidelity spectra to influence the training pipeline, whereas an overly restrictive threshold can unnecessarily discard useful high-quality samples and limit acquisition diversity, ultimately reducing learning effectiveness.}

\rev{Table~\ref{tab:tau_sensitivity} summarizes the final three acquisition steps for both Im2Spec and Spec2Im. Across both tasks, the results show that ActiveQC is sensitive to the choice of $\tau$, but that moderate threshold values that effectively separate low-quality measurements from high-quality data provide stable and strong performance. In the main simulated benchmark experiments, we used $\tau = 0.30$, which was selected empirically by the domain expert based on the observed spectral quality and the desired balance between suppressing low-fidelity measurements and preserving useful spectral diversity. The sensitivity analysis further confirms that the selected threshold lies within the intermediate regime that provides near-optimal performance across both tasks and highlights the importance of human-in-the-loop quality-threshold selection in balancing data-fidelity enforcement with sufficient acquisition diversity.}
\section*{\rev{S3. Step-by-Step Evolution of Quality-Controlled Acquisition During Real-Time AFM Deployment}}

\begin{figure*}[htbp]
    \centering
    \includegraphics[width=0.90\textwidth]{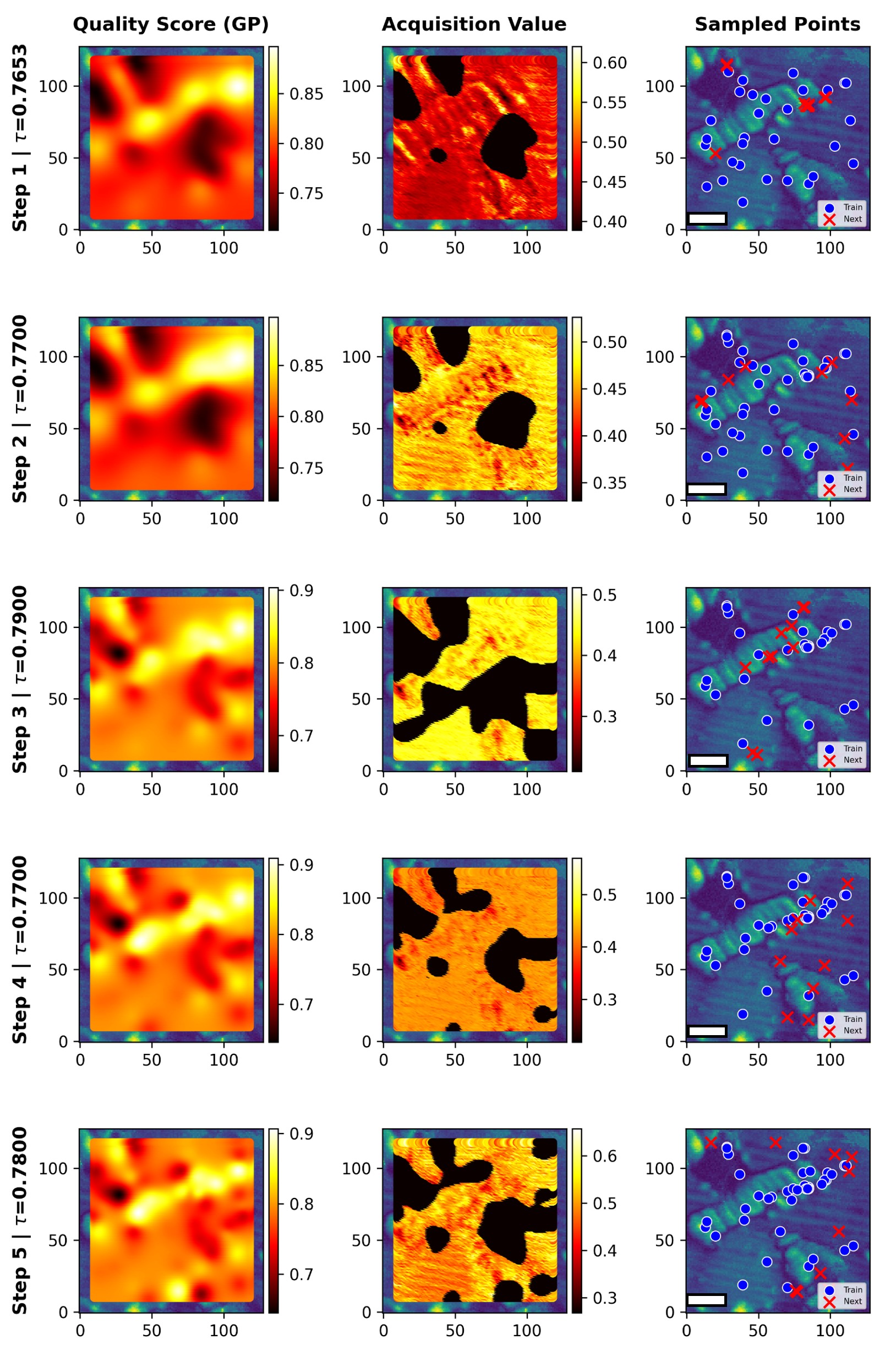}
    \caption{\rev{Step-by-step ActiveQC acquisition for PTO Region~1, exploration Steps~1--5. For each step, columns show the GP-predicted quality score, final acquisition-value map after quality-based gating, and sampled points. Blue circles indicate retained training samples and red crosses indicate newly selected acquisition points. The quality threshold $\tau$ used at each step is shown in the row label. Scale bars correspond to 400~nm.}}
    \label{fig:si_rti_region1_steps_1_5}
\end{figure*}

\begin{figure*}[htbp]
    \centering
    \includegraphics[width=0.90\textwidth]{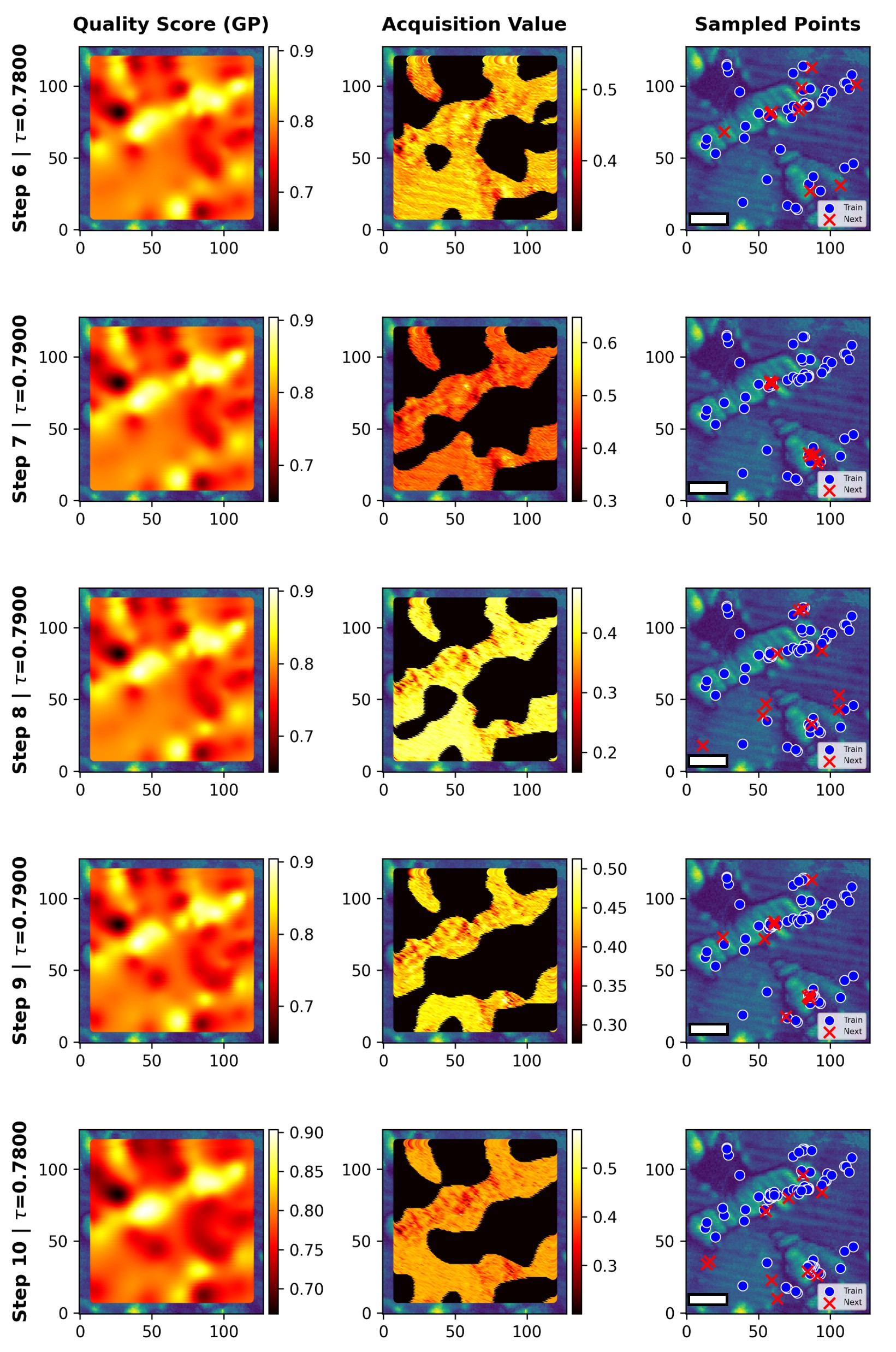}
    \caption{\rev{Step-by-step ActiveQC acquisition for PTO Region~1, exploration Steps~6--10. Same layout as Fig.~\ref{fig:si_rti_region1_steps_1_5}, showing the continued evolution of GP-predicted quality scores, gated acquisition values, and selected acquisition points. Scale bars correspond to 400~nm.}}
    \label{fig:si_rti_region1_steps_6_10}
\end{figure*}

\begin{figure*}[htbp]
    \centering
    \includegraphics[width=0.90\textwidth]{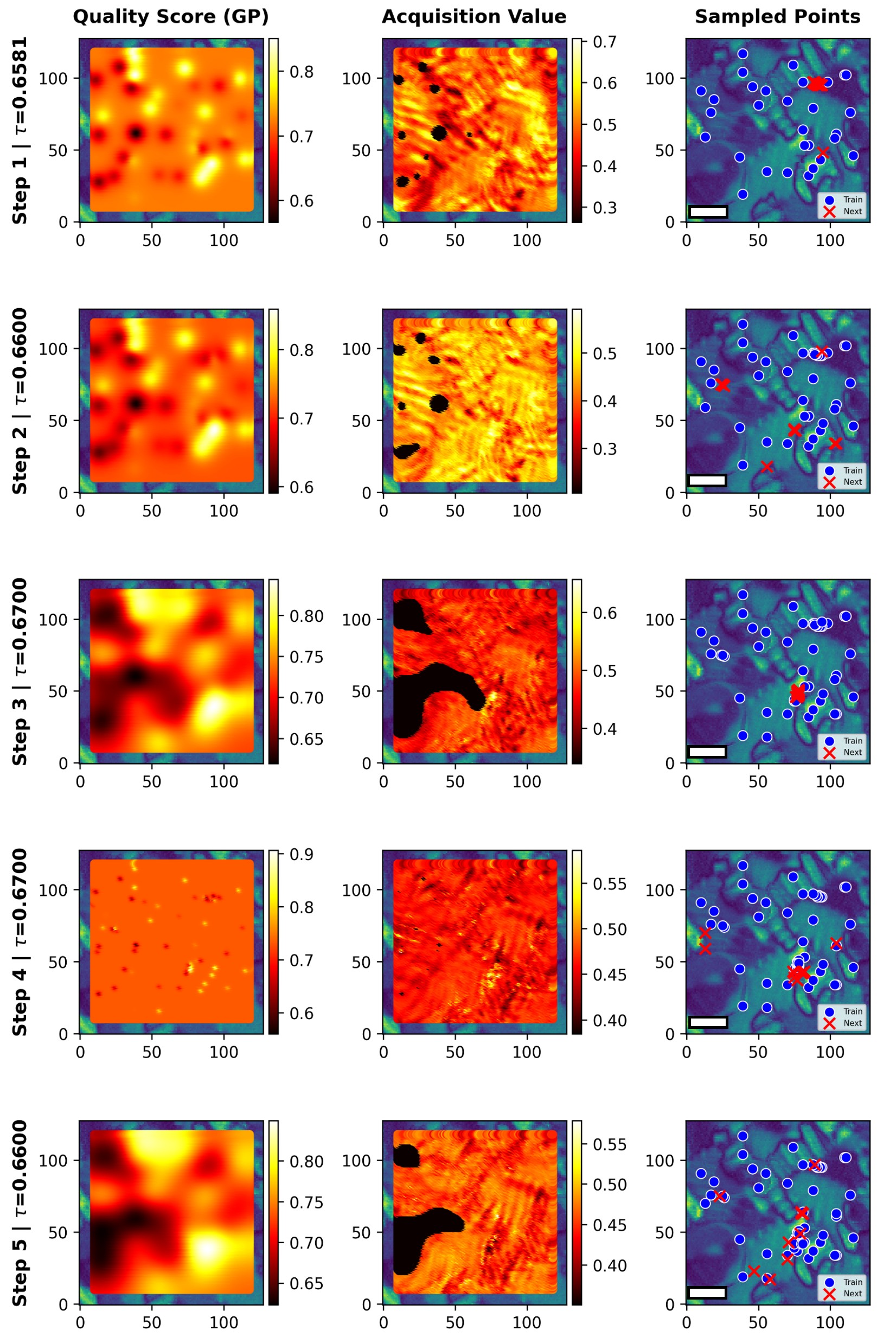}
    \caption{\rev{Step-by-step ActiveQC acquisition for PTO Region~2, exploration Steps~1--5. For each step, columns show the GP-predicted quality score, final acquisition-value map after quality-based gating, and sampled points. Blue circles indicate retained training samples and red crosses indicate newly selected acquisition points. The quality threshold $\tau$ used at each step is shown in the row label. Scale bars correspond to 400~nm.}}
    \label{fig:si_rti_region2_steps_1_5}
\end{figure*}

\begin{figure*}[htbp]
    \centering
    \includegraphics[width=0.90\textwidth]{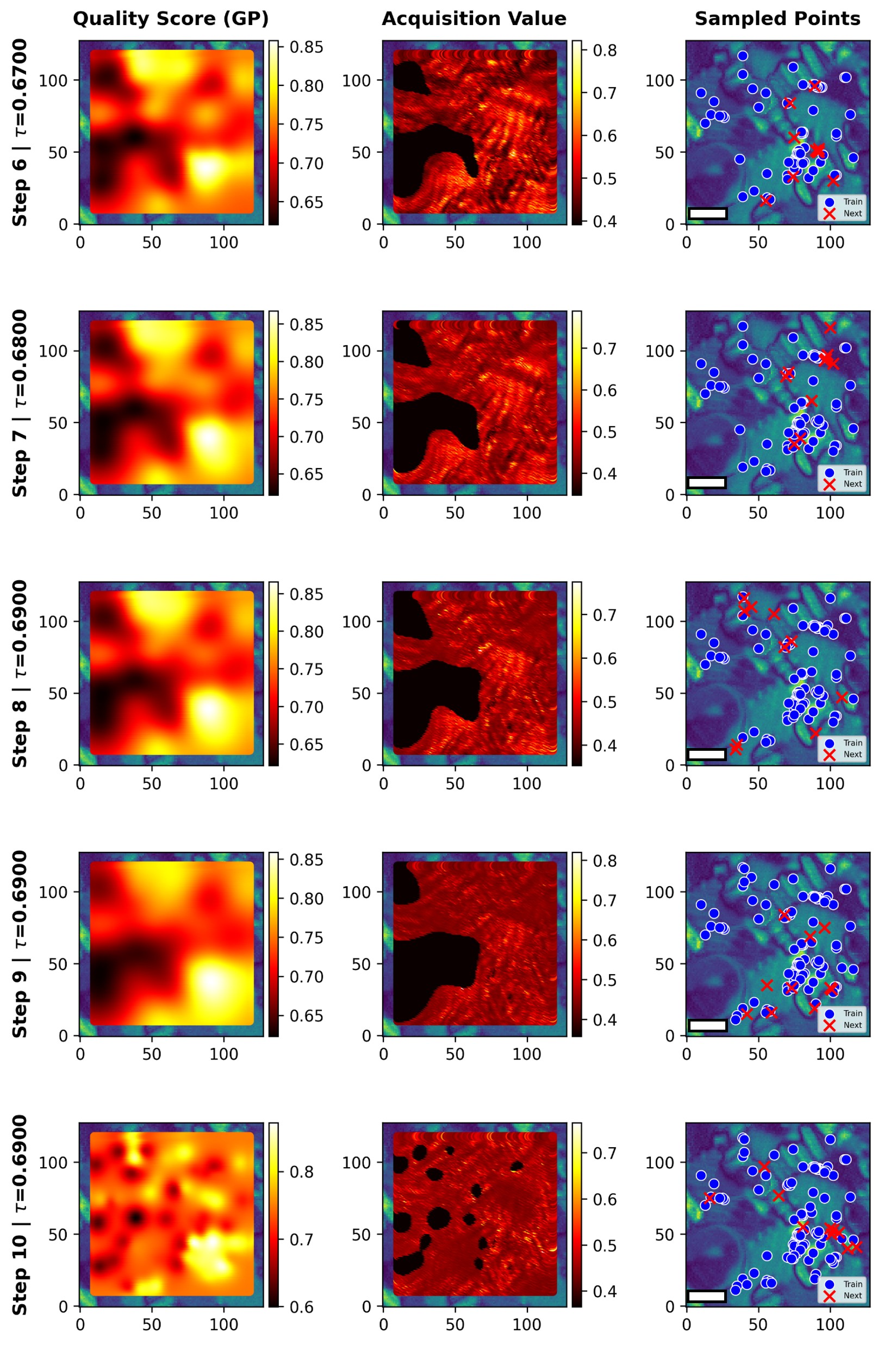}
    \caption{\rev{Step-by-step ActiveQC acquisition for PTO Region~2, exploration Steps~6--10. Same layout as Fig.~\ref{fig:si_rti_region2_steps_1_5}, showing the continued evolution of GP-predicted quality scores, gated acquisition values, and selected acquisition points. Scale bars correspond to 400~nm.}}
    \label{fig:si_rti_region2_steps_6_10}
\end{figure*}

\begin{figure*}[htbp]
    \centering
    \includegraphics[width=0.90\textwidth]{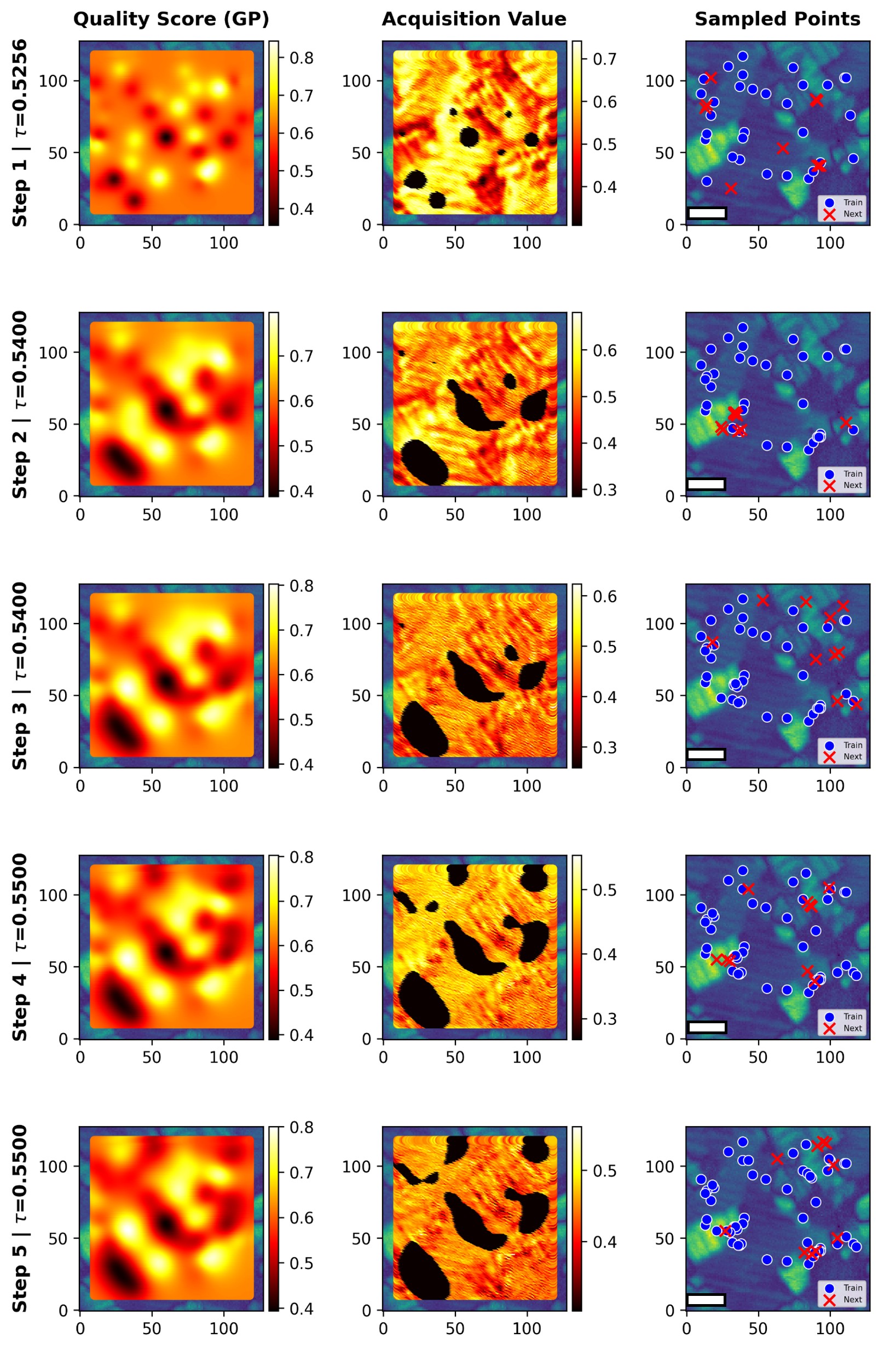}
    \caption{\rev{Step-by-step ActiveQC acquisition for PTO Region~3, exploration Steps~1--5. For each step, columns show the GP-predicted quality score, final acquisition-value map after quality-based gating, and sampled points. Blue circles indicate retained training samples and red crosses indicate newly selected acquisition points. The quality threshold $\tau$ used at each step is shown in the row label. Scale bars correspond to 400~nm.}}
    \label{fig:si_rti_region3_steps_1_5}
\end{figure*}

\begin{figure*}[htbp]
    \centering
    \includegraphics[width=0.90\textwidth]{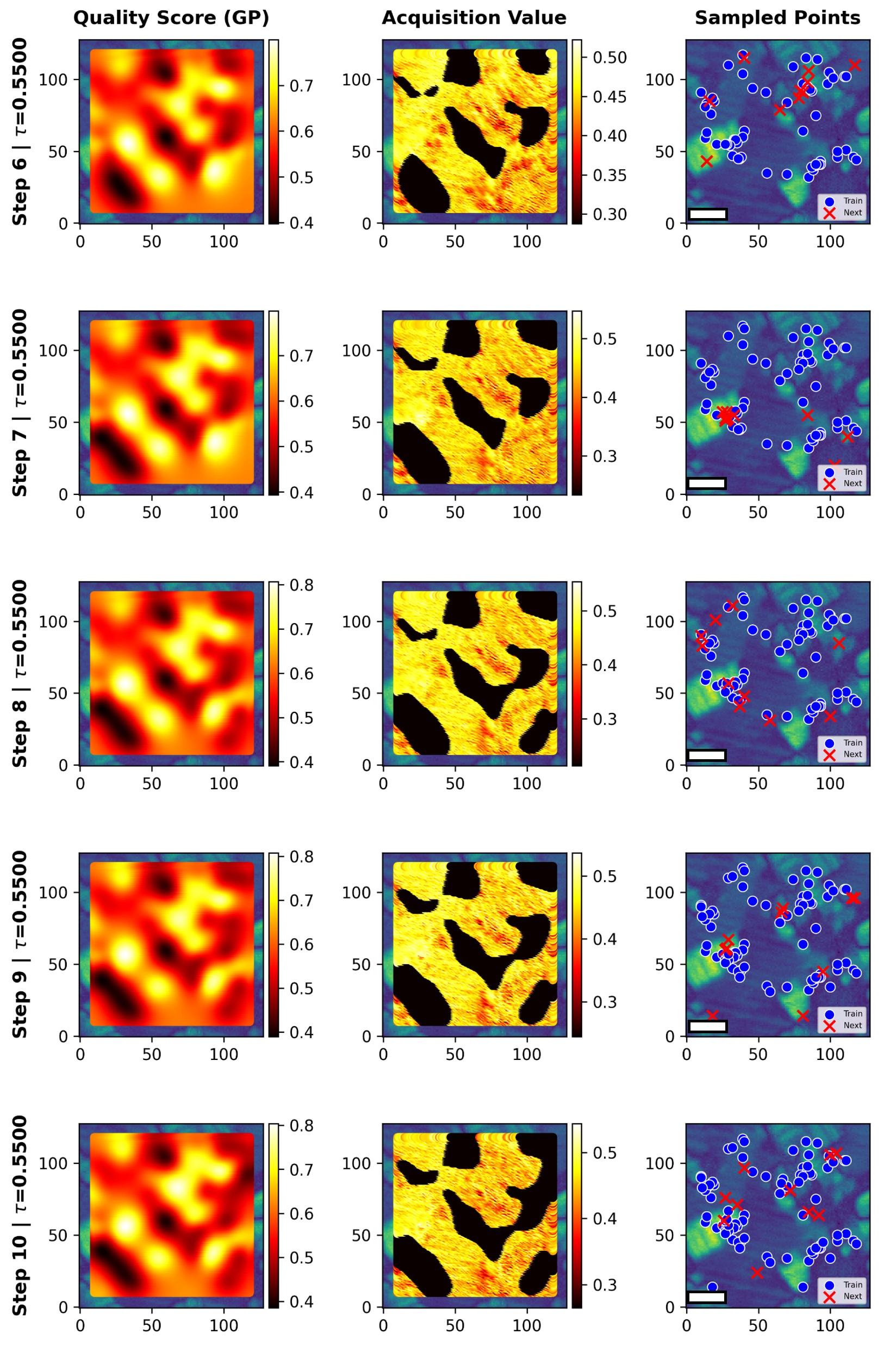}
    \caption{\rev{Step-by-step ActiveQC acquisition for PTO Region~3, exploration Steps~6--10. Same layout as Fig.~\ref{fig:si_rti_region3_steps_1_5}, showing the continued evolution of GP-predicted quality scores, gated acquisition values, and selected acquisition points. Scale bars correspond to 400~nm.}}
    \label{fig:si_rti_region3_steps_6_10}
\end{figure*}

\rev{Figures~\ref{fig:si_rti_region1_steps_1_5}--\ref{fig:si_rti_region3_steps_6_10} show the step-by-step evolution of the quality-controlled active learning process during real-time AFM deployment on three independent PTO regions explored using ActiveQC. For each region, the 10 exploration steps are grouped into two composite figures: Steps~1--5 and Steps~6--10.}

\rev{At each exploration step, three quantities are visualized on the same structural PFM background image: (i) the GP-predicted spectral quality score, (ii) the final acquisition-value map after applying quality-based gating, and (iii) the retained training samples together with newly selected acquisition points. Blue circles denote the retained training samples available at that step, while red crosses indicate the next points selected for acquisition. The quality threshold $\tau$ used at each step is reported along the corresponding row label. These visualizations show how the GP-based quality model and threshold-dependent gating jointly shape the acquisition landscape over successive exploration steps.}

\rev{Across the three regions, the acquisition maps demonstrate that ActiveQC dynamically suppresses areas predicted to have lower spectral quality while preserving candidate locations with higher acquisition utility. Depending on the quality threshold $\tau$ selected at a given exploration step, the degree of suppression also changes. The sampled-point maps further show that the resulting acquisitions concentrate around spatially informative regions such as near domain walls, where more complex electromechanical responses are expected.}

